\documentclass[5p,twocolumn]{elsarticle}

\usepackage{hyperref}
\usepackage{lineno}
\usepackage{times}
\usepackage{epsfig}
\usepackage{amsmath,amsfonts,amsthm,bm}
\usepackage{amssymb}
\usepackage{graphicx}
\usepackage{epstopdf}
\usepackage[normal]{subfigure}
\usepackage{verbatim}
\usepackage{picins}
\usepackage{diagbox}
\usepackage{enumerate}
\usepackage{algorithm}
\usepackage{algpseudocode}
\usepackage{booktabs}
\usepackage[table]{xcolor} 
\usepackage[group-separator={,}]{siunitx}
\usepackage{makecell}
\usepackage{siunitx}
\usepackage{url}
\usepackage{multirow}
\usepackage{amssymb}
\usepackage{pifont}

\usepackage{xspace}
\usepackage{tikz}

\biboptions{numbers,sort&compress}
\makeatletter
\DeclareRobustCommand\onedot{\futurelet\@let@token\@onedot}
\def\@onedot{\ifx\@let@token.\else.\null\fi\xspace}

\modulolinenumbers[5]

\def\R #1|#2|#3|#4{ #1&#2&#3&#4}
\journal{Journal of Neurocomputing}

\usepackage[bottom]{footmisc}

\usepackage[export]{adjustbox}

\usepackage{expl3}

\newcommand{\etal}{\textit{et al}. }

\newcommand{\eg}{\textit{e}.\textit{g}. }
\def\etc{etc.\@\xspace}

\usepackage[]{caption2}

\begin{document}
\begin{frontmatter}
\title{AU R-CNN: Encoding Expert Prior Knowledge into R-CNN for Action Unit Detection}

\author[mymainaddress]{Chen Ma}

\ead{sharpstill@163.com}
\author[mymainaddress]{Li Chen\corref{mycorrespondingauthor}}
\ead{chenlee@tsinghua.edu.cn}
\author[mymainaddress]{Junhai Yong}
\ead{yongjh@tsinghua.edu.cn}

\cortext[mycorrespondingauthor]{Corresponding author: Li Chen}

\address[mymainaddress]{School of Software, Tsinghua University, Beijing 100084, China\\Beijing National Research Center for Information Science and Technology (BNRist)}

\begin{abstract}
	Detecting action units (AUs) on human faces is challenging because various AUs make subtle facial appearance change over various regions at different scales. Current works have attempted to recognize AUs by emphasizing important regions. However, the incorporation of expert prior knowledge into region definition remains under-exploited, and current AU detection approaches do not use regional convolutional neural networks (R-CNN) with expert prior knowledge to directly focus on AU-related regions adaptively.
	By incorporating expert prior knowledge, we propose a novel R-CNN based model named AU R-CNN. The proposed solution offers two main contributions: (1) 
    AU R-CNN directly observes different facial regions, where various AUs are located. Specifically, we define an AU partition rule which encodes the expert prior knowledge into the region definition and RoI-level label definition. This design produces considerably better detection performance than existing approaches. (2) We integrate various dynamic models (including convolutional long short-term memory, two stream network, conditional random field, and temporal action localization network) into AU R-CNN and then investigate and analyze the reason behind the performance of dynamic models. Experiment results demonstrate that \textit{only} static RGB image information and no optical flow-based AU R-CNN surpasses the one fused with dynamic models. AU R-CNN is also superior to traditional CNNs that use the same backbone on varying image resolutions. State-of-the-art recognition performance of AU detection is achieved. 
    The complete network is end-to-end trainable. Experiments on BP4D and DISFA datasets show the effectiveness of our approach. The implementation code is available in \url{https://github.com/sharpstill/AU_R-CNN}.
\end{abstract}

\end{frontmatter}

\section{Introduction}
\begin{figure*}[htbp]
	\includegraphics[width=\textwidth,scale=0.8]{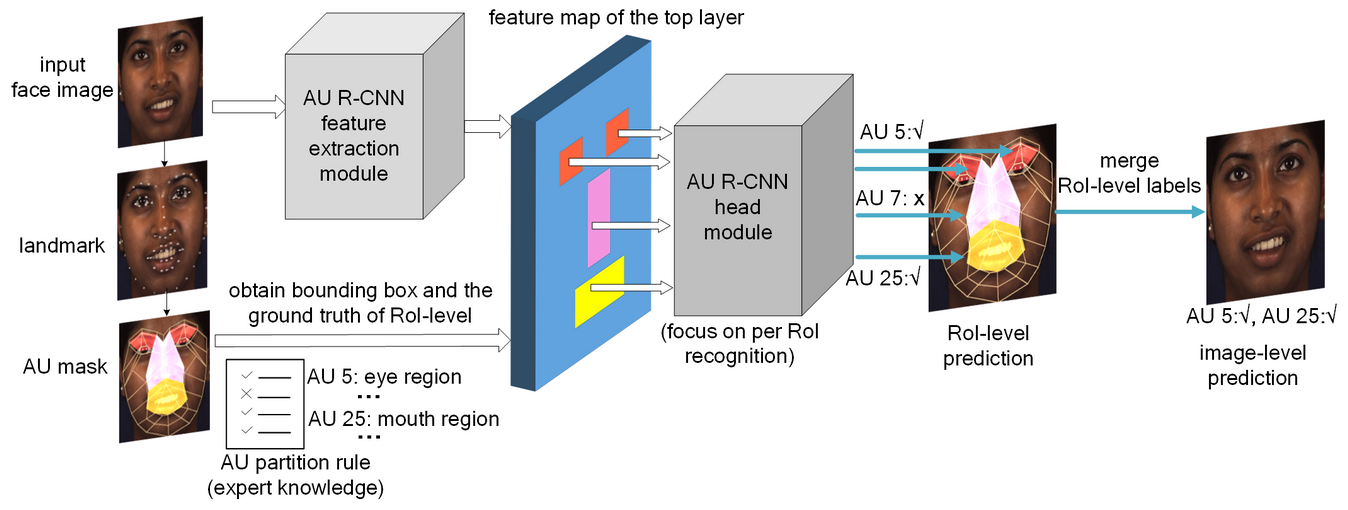}
	\caption{The overall of AU R-CNN framework. It recognizes each RoI-level's label based on the AU partition rule, which uses the landmark point location information to encode the expert prior knowledge. This rule indicates the place where the AUs may occur. AU R-CNN head module focuses on recognizing each bounding box to improve performance.}
	\label{fig:tease}
\end{figure*}
Facial expressions reveal people's emotions and intentions. Facial Action Coding System (FACS) \cite{ekman1977facial} has defined 44 action units (AUs) related to the movement of specific facial muscles; these units can anatomically represent all possible facial expressions, considering the crucial importance of facial expression analysis. AU detection has been studied for decades and its goal is to recognize and predict AU labels on each frame of the facial expression video. Automatic detection of AUs has a wide range of applications, such as human-machine interfaces, affective computing, and car-driving monitoring.

Since the human face may present complex facial expression, and AUs appear in the form of subtle appearance changes on the local regions of face, that current classifiers cannot easily recognize. This problem is the main obstacle of current AU detection systems. 
Various approaches focus on fusing with extra information in convolutional neural networks (CNNs), \eg, the optical flow information \cite{romero2017multi} or landmark information \cite{li2017action,li2017eac}, to help AU detection systems capture such subtle facial expressions. However, these approaches have high detection error rates, due to the lack of using prior knowledge. Human can easily recognize micro facial expression by their long accumulated experience. Hence, integrating the expert prior knowledge of FACS \cite{ekman1977facial} to AU detection system is promising. With fusing of this prior knowledge, our proposed approach addresses the AU detection problem by partitioning the face to easily recognizable AU-related regions, then the prediction of each region is merged to obtain the image-level prediction. Fig. \ref{fig:tease} shows our approach's framework, we design an ``AU partition rule" to encode the expert prior knowledge. This AU partition rule decomposes the image into a bunch of AU-related bounding boxes. Then AU R-CNN head module focuses on recognizing each bounding box. This design can well address the three problems of existing approaches.

First, existing approaches \cite{eleftheriadis2015multi,koelstra2010dynamic,wang2013capturing,chu2013selective,ding2013facial,zeng2015confidence,liu2013aware,valstar2015fera,Zhao2016,li2017eac,li2017action} have been proposed to extract features near landmarks (namely, ``AU center"), which is trivially defined and leading to emphasize on inaccurate places. AUs occur in regions around specific facial muscles that may be inaccurately located on a landmark or an AU center due to the limitation of the facial muscle's activity place. Thus, most AUs limit their activities in specific irregular regions of a face, and we call this limitation the ``space constraint". Our approach reviews the FACS and designs the ``AU partition rule" to represent this space constraint accurately. This well-designed ``AU partition rule" is called the ``expert prior knowledge" in our approach which is built on the basis of the space-constraint for regional recognition, so it reduces the detection error rate caused by inaccurate landmark positioning (see experiment Section \ref{sec:mean_box}).

Second, existing approaches still use CNNs to recognize a full face image \cite{Zhao2016b,li2017eac,li2017action,han2017optimizing} and do not learn to recognize individual region's labels, which may not use the correct image context to detect. For example, a CNN may use an unreliable context, such as mouth area features, to recognize eye-area-related AUs (\eg AU 1, AU 2). Recent success in the object detection model of Fast/Faster R-CNN \cite{girshick2015fast,Ren2015} has inspired us to utilize the power of R-CNN based models to learn the accurate regional features of AUs under space constraints. We propose AU R-CNN to detect AUs only from AU-related regions by limiting its vision inside space-constrained areas. In this process, unrelated areas can be excluded to avoid interference, which is key to improve detection accuracy.

Third, the multi-label learning problem in AU detection can be addressed at a fine-grained level under AU-related RoI space constraint. Previous approaches \cite{Zhao2016b,li2017action,li2017eac} adopt the sigmoid cross-entropy cost function to learn the image-level multi-label and emphasize the important regions, but such a solution is not sufficiently fine-grained. The multi-label relationship can be captured more accurately in the RoI-level supervised information constraint. Most facial muscles can show diverse expressions that lead to RoI-level multi-label learning. For example, AU 12 (lip corner puller) is often present in a smile, which may also occur together with AU 10 (upper lip raiser), and deepen the nasolabial fold, as shown in Fig. \ref{fig:ROI_face}. Therefore, in the definition of the AU partition rule, AUs are grouped by the definition of FACS and related facial muscles. Each AU group shares the same region, and such AU group can be represented by a binary vector with element of 1 if the corresponding AU occurs in the ground truth and 0 otherwise. The sigmoid cross-entropy cost function is adopted in the RoI-level learning. In our experiments, we determine that using RoI-level labels to train and predict and then merging the RoI-level prediction result to that of the image level surpasses the previous approaches.

Furthermore, we analyze the effects of fusing temporal features into AU R-CNN (dynamic model extension). We conduct complete comparison experiments to investigate the effects of integrating dynamic models, including convolutional long short-term memory (ConvLSTM) \cite{xingjian2015convolutional}, two-stream network \cite{feichtenhofer2016convolutional}, general graph conditional random field (CRF) model, and TAL-Net \cite{chao2018rethinking}, into AU R-CNN. We analyze the reason behind such effects and the cases under which the dynamic models are effective. Our AU R-CNN with \textit{only} static RGB images and no optical flow achieves 63\% average F1 score on BP4D, and outperforms all dynamic models.
The main contributions of our study are as follows.
  
  (1) AU R-CNN is proposed to learn regional features adaptively by using RoI-level multi-label supervised information. Specifically, we encode the expert prior knowledge by defining the AU partition rule, including the AU groups and related regions, according to FACS \cite{ekman1977facial}.
  
  (2) We investigate the effects of integrating various dynamic models, including two-stream network, ConvLSTM, CRF model and TAL-Net, in the experiments of BP4D \cite{zhang2014bp4d} and DISFA \cite{mavadati2013disfa} databases. The reasons behind such experiment effects and the effective cases are analyzed. The experiment results show that our static RGB image-based AU R-CNN achieves the best average F1 score in BP4D and is close to the performance of the best dynamic model in DISFA. Our approach achieves state-of-the-art performance in AU detection.

\section{Related Work}

Extensive works on AU detection have been proposed to extract effective facial features. The facial features in AU detection can be grouped into appearance and geometric features. Appearance features portray the local or global changes in facial components. Most popular approaches in this category adopt Haar feature \cite{whitehill2006haar}, local binary pattern \cite{jiang2011action}, Garbor wavelets \cite{bazzo2004recognizing,valstar2006fully}, and canonical appearance feature \cite{lucey2010extended}. Geometric features represent the salient facial point or skin changing direction or distance. Geometric changes can be measured by optical flows \cite{lien2000detection} or displacement of facial landmark points \cite{valstar2012fully,lucey2010extended}. Landmark plays an important role in geometry approaches, and many methods have been proposed to extract features near landmark points \cite{eleftheriadis2015multi,koelstra2010dynamic,wang2013capturing,chu2013selective,ding2013facial,zeng2015confidence,liu2013aware,valstar2015fera, yu2018rethining}. Fabian \etal \cite{fabian2016emotionet} proposed a method that combines geometric changes and local texture information. Wu and Ji \cite{Wu2016} investigated the combination of facial AU recognition and facial landmark detection. Zhao \etal \cite{Zhao2016} proposed joint patch and multi-label learning (JPML) for AU detection with a scale-invariant feature transform descriptor near landmarks. These traditional approaches focus on extracting handcraft features near landmark points. With the recent success of deep learning, CNN has been widely adopted to extract AU features \cite{han2017optimizing}. Zhao \etal \cite{Zhao2016b} proposed a deep region and multi-label learning (DRML) network to divide the face images into $8\times 8$ blocks and used individual convolutional kernels to convolve each block. Although this approach treats each face as a group of individual parts, it divides blocks uniformly and does not consider the FACS knowledge, thereby leading to the poor performance.
Li \etal \cite{li2017eac} proposed Enhancing and Cropping Net (EAC-Net), which intends to give significant attention to individual AU centers. However, this approach defines the AU center trivially and it uses image-level context to learn. Its CNN backbone may use incorrect context to classify and the lack of RoI-level supervised information can only give coarse guidance. Song \etal \cite{song2015exploiting} studied the sparsity and co-occurrence of AUs. Han \etal \cite{han2017optimizing} proposed an Optimized Filter Size CNN (OFS-CNN) to simultaneously learn the filter sizes and weights of all conv-layer.
Other related problems, including the effects of dataset size \cite{girard2015much}, the action detection in videos \cite{hou2017tube}, the pose-based feature of action recognition \cite{cheron2015p}, and generalized multimodal factorized high-order pooling for visual question answering \cite{yu2018beyond} have also been studied. Previous works have mainly focused on landmark-based regions or learning multiple regions with convolutional kernels separately. Detection with the expert prior knowledge and utilizing RoI-level labels are important but have been undervalued in previous methods.

Researchers have utilized temporal dependencies in video sequences over the last few years. Romero \etal \cite{romero2017multi} advocated a two-stream CNN model that combines optical flow and RGB information, and their result was promising. However, they used one binary classification model for each AU, which caused their approach to be time consuming to train and yield numerous model parameters. The CNN and LSTM hybrid network architectures are studied in Chu \etal \cite{Chu2017Learning}, Li \etal \cite{li2017action} and He \etal \cite{he2017multi}, which feed the CNN-produced features to LSTM to improve performance by capturing the temporal relationship across frames. However, their solutions are inefficient because they are not an end-to-end networks.
In our experiments, we also investigate the effects of using temporal feature relationships in the time axis of videos. We use various dynamic models (including two-stream network, ConvLSTM \etc) that are incorporated into AU R-CNN. Such temporal dependency cannot always improve performance in all cases (Section \ref{sec:static_vs_dynamic}).

Unlike existing approaches, AU R-CNN is a unified end-to-end learning model that encodes expert prior knowledge and outperforms state-of-the-art approaches. Thus, it is a simple and practical model.

\section{Proposed Method}

\subsection{Overview}
AU detection can be considered a multi-label classification problem. The most popular image classification approach is the CNN, and the basic assumption for a standard CNN is the shared convolutional kernels for an entire image. For a highly structural image, such as a human face, a standard CNN will fail to capture subtle appearance changes. To address this issue, we propose AU R-CNN, in which expert prior knowledge is encoded. We review FACS \cite{ekman1977facial} and define a rule (``AU partition rule") for partitioning a face on the basis of FACS knowledge using landmarks. With this rule, we can treat each face image as a group of separate regions and AU R-CNN is proposed to recognize each region. The overall procedure is composed of two steps. First, the face image's landmark points are obtained, and then the face is partitioned into regions on the basis of the AU partition rule and the landmark coordinates. The ``AU masks" are generated in this step, and the expert prior knowledge is encoded into the AU masks. Second, the face images are input into the AU R-CNN's backbone, the produced feature map and the minimum bounding boxes of the AU mask are then fed into AU R-CNN's RoI pooling layer together. The final fully-connected (fc) layer's output can be treated as classification probabilities. The image-level ground truth label is also partitioned to RoI-level in the learning. After AU R-CNN is trained over, the prediction is performed on the RoI-level. Then, we use a ``bit-wise OR" operator to merge RoI-level prediction labels to image-level ones. 
In this section, we introduce the AU partition rule and then AU R-CNN. We also introduce a dynamic model extension of AU R-CNN in Section \ref{sec:method_AR_conv-lstm}.

\begin{table}
	\small	
	\centering
	\caption{FACS definition of AUs and related muscles \cite{ekman1977facial}}
	\label{tab:FACS}
	\tabcolsep=0.05cm
	\begin{tabular}{ccc}
		\toprule
		AU number &  AU name & Muscle Basis \\
		\midrule
		1 & Inner brow raiser & Frontalis \\
		2 & Outer brow raiser & Frontalis \\
		4 & Brow lowerer & Corrugator supercilii \\
		6 & Cheek raiser & Orbicularis oculi \\
		7 & Lid tightener & Orbicularis oculi \\
		10 & Upper lip raiser & Levator labii superioris \\
		12 & Lip corner puller & Zygomaticus major \\
		14 & Dimpler & Buccinator \\
		15 & Lip corner depressor &  Depressor anguli oris  \\
		17 & Chin raiser & Mentalis \\
		23 & Lip tightener & Orbicularis oris \\
		24 & Lip pressor & Orbicularis oris \\
		25 & Lips part	& Depressor labii inferioris\\
		26 & Jaw drop & Masseter \\	
		\bottomrule
	\end{tabular}
	\vspace{-0.5cm}
\end{table}

 \subsection{AU partition rule}
\label{sec:AU partition rule}

\begin{figure}[htbp]
	\begin{adjustbox}{max width=0.5\textwidth,center}
		\subfigure{
			\includegraphics[width=0.25\textwidth,scale=0.8]{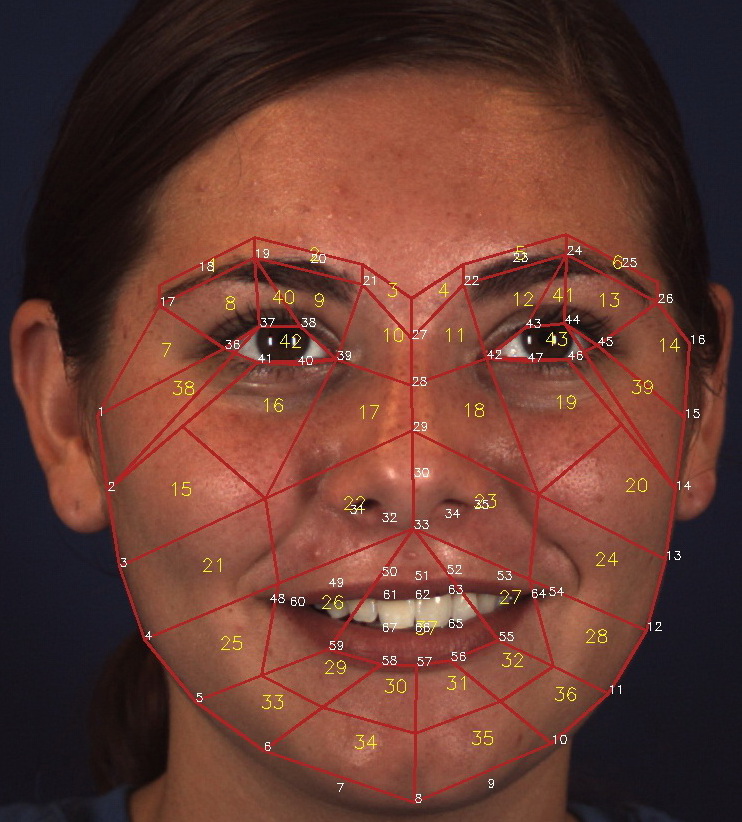}
		}
		\subfigure{
			\label{fig:face_detail}
			\includegraphics[width=0.25\textwidth]{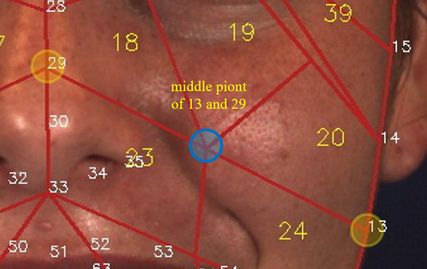}	
		}
	\end{adjustbox}
	\caption{Landmark and region partition of face. Yellow and white numbers indicate the RoI number and landmark number respectively. \textbf{Left:} Partition of $43$ RoIs. \textbf{Right:} Position of blue point is the average position of landmark $13$ and $29$.}
	\label{fig:ROI_face}
\end{figure}

AUs appear in specific regions of a face but are not limited to facial landmark points; previous AU feature extraction approaches directly use facial landmarks or offsets of the landmarks as AU centers \cite{li2017eac,Zhao2016,li2017action}, but the actual places where activities occur may be missed, and sensitivity of the system may be increased. Instead of identifying the AU center, we adopt the domain-related expertise to guide the partition of AU-related RoIs. The first step is to utilize the dlib \cite{king2009dlib} toolkit to obtain 68 landmark points. The landmark points provide rich information about the face, and the landmark points help us focus on areas where AUs may occur. Fig. \ref{fig:ROI_face} shows the region partition of a face, and several extra points are calculated using 68 landmarks. A typical example is shown in Fig. \ref{fig:ROI_face} right. The face image is partitioned into 43 basic RoIs using landmarks. Then, on the basis of FACS definition\footnote{\url{https://www.cs.cmu.edu/~face/facs.htm}} (Table \ref{tab:FACS}) and the anatomy of facial muscle structure\footnote{\url{https://en.wikipedia.org/wiki/Facial_muscles}}, the AU partition rule and the AU mask can be defined for representing the expert prior knowledge. For this purpose, we classify AUs into four cases.

(1) The RoIs defined in Fig. \ref{fig:ROI_face} are the basic building blocks, named basic RoIs. One AU contains multiple basic RoIs; hence, multiple basic RoIs are selected to be grouped and assigned to AUs by RoI numbers (Table \ref{tab:AU_region}). The principle of such RoI assignment is the FACS muscle definition (Table \ref{tab:FACS}). The region of the grouped RoIs is called the ``AU mask".

\begin{figure}[h]
	\begin{adjustbox}{max width=0.5\textwidth,center}
		\subfigure[AU group \# $1$]{
			\label{fig:mask_1}
			\includegraphics[width=0.11\textwidth]{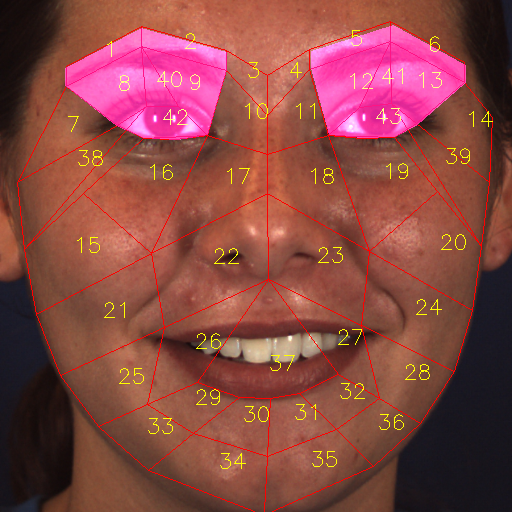}
		}
		\subfigure[AU group \# $2$]{
			\label{fig:mask_2}
			\includegraphics[width=0.11\textwidth]{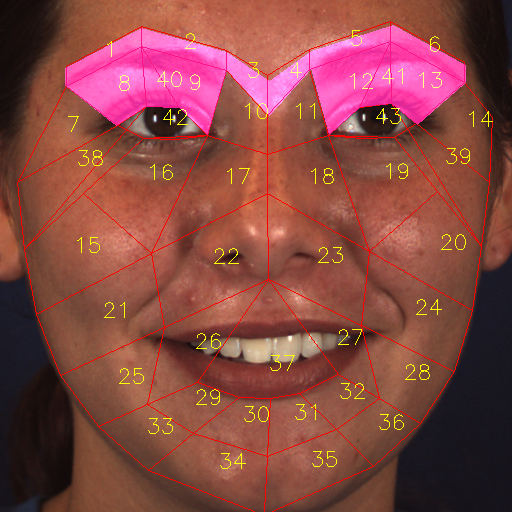}
		}
		\subfigure[AU group \# $3$]{
			\label{fig:mask_3}
			\includegraphics[width=0.11\textwidth]{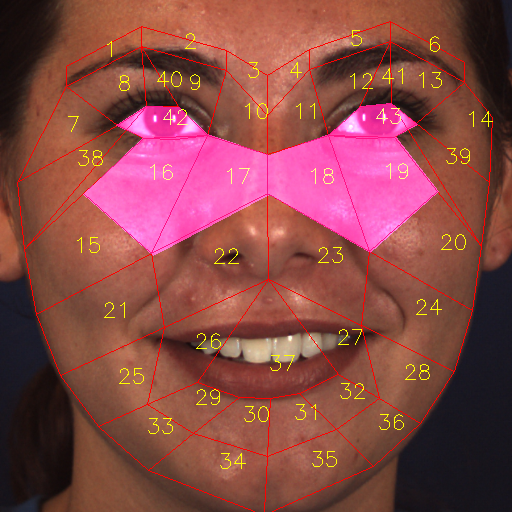}
		}
		\subfigure[AU group \# $4$]{
			\label{fig:mask_4}
			\includegraphics[width=0.11\textwidth]{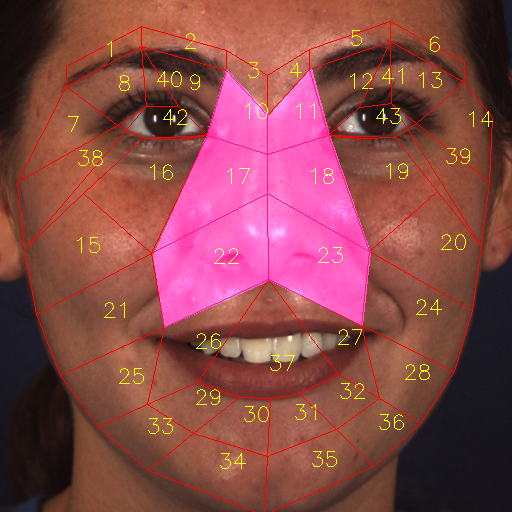}
		}
	\end{adjustbox}
	\begin{adjustbox}{max width=0.5\textwidth,center}
		
		\subfigure[AU group \# $5$]{
			\label{fig:mask_5}
			\includegraphics[width=0.11\textwidth]{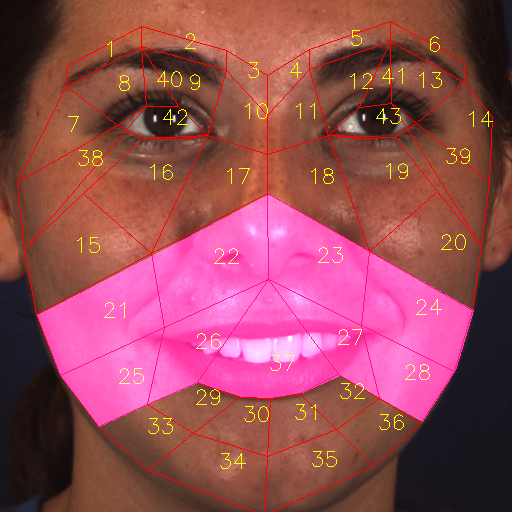}
		}
		\subfigure[AU group \# $6$]{
			\label{fig:mask_6}
			\includegraphics[width=0.11\textwidth]{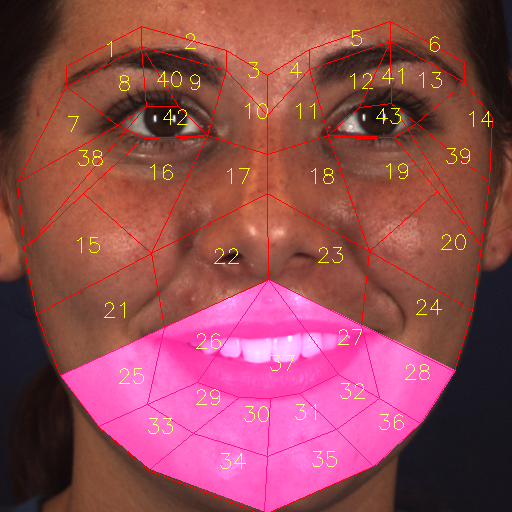}
		}
		\subfigure[AU group \# $7$]{
			\label{fig:mask_7}
			\includegraphics[width=0.11\textwidth]{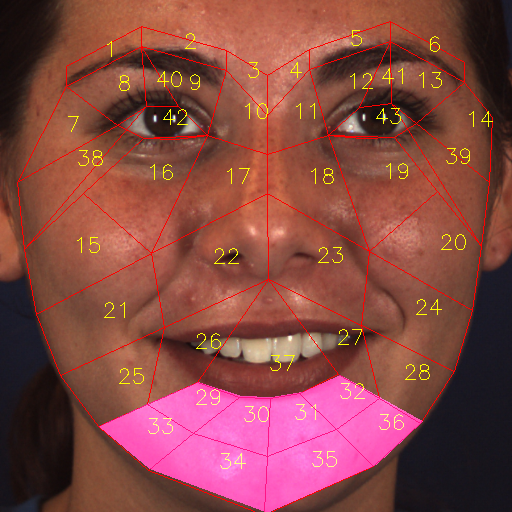}
		}
		\subfigure[AU group \# $8$]{
			\label{fig:mask_8}
			\includegraphics[width=0.11\textwidth]{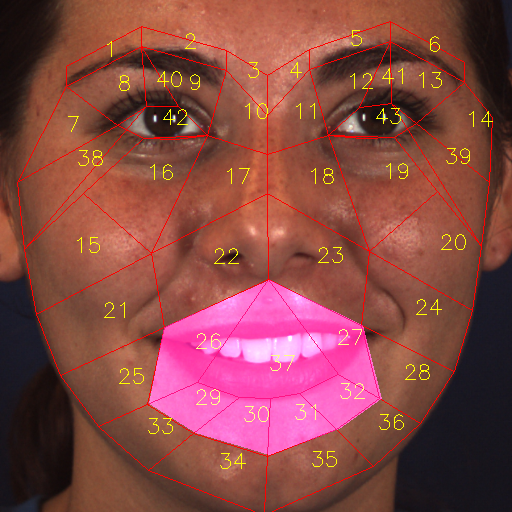}
		}
	\end{adjustbox}
	
	\caption{Action Unit masks for AU group \#$1$ $\sim$ \#$8$ (see Table \ref{tab:AU_region}).}
	\label{fig:AU_mask}
\end{figure}

(2) Most muscles can present multiple AUs---in other words, some AUs can co-occur in the same place. For example, AU 12 (lip corner puller) and AU 10 (upper lip raiser) are often present together in a smile, which requires lifting of the muscle and may also deepen the nasolabial fold, as shown in Fig. \ref{fig:mask_5}. Therefore, we group AUs into 8 ``AU groups" on the basis of AU-related muscles defined in FACS (Table \ref{tab:FACS}) and the AU co-occurrence statistics of the database. Each AU group has its own mask, whose region is shared by the AUs. One AU group contains multiple basic RoIs, which are defined in Fig. \ref{fig:ROI_face}, to form an AU mask (Fig. \ref{fig:AU_mask}). 

(3) Some AU groups are defined in a hierarchical structure, that is, these AU group masks have a broad area, which may contain other AU groups' small areas. For example, AU group \# 6 contains AU group \# 7 (Fig. \ref{fig:AU_mask}). The reason behind such a design is that AU group \# 7 (AU 17) is caused by the movement of the mentalis (Table \ref{tab:FACS}), which is in the chin. The bone structure of the chin makes it a relatively stable area, which limits the possible occurrence of AU 17. Therefore, we can define a detailed area in AU group \# 7 (Fig.\ref{fig:mask_7}). However, AU group \# 6 consists of AU 16, AU 20, AU 25 and AU 26, and it is located in the mouth area. The mouth area contains several possible movement locations (mouth open, mouth close, smell, laugh, \etc), and the chin area follows mouth opening and closing. Therefore we define AU group \# 6 to contain the area of AU group \# 7 (Fig. \ref{fig:mask_6}). The partition of the face image naturally leads to the RoI-level label assignment. In this case, the AU group \# 6 must contain RoI-level labels of AU group \# 7. We define operator ``label fetch" \#7$ \in $\#6 to enable AU group \# 6 to fetch labels from AU group \#7 (Table \ref{tab:AU_region}).

(4) Some AU groups have overlapping areas with other AU groups' areas. For example, AU group \# 3's mask, which is across the nose area (Fig. \ref{fig:mask_3}), will also contain labels of AU group \# 4 (Fig. \ref{fig:mask_4}); thus, we also use the operator ``label fetch" \#4$\in $\#3 to fetch labels from AU group \# 4 in this case. (Table \ref{tab:AU_region}).

In summary, Table \ref{tab:AU_region} and Fig. \ref{fig:AU_mask} show the AU partition rule and the AU mask. The AU group definition is related not only to the RoI partition of the face, but also to the RoI-level label assignment.

	\renewcommand{\arraystretch}{0.5} 
\begin{table}[!t]

	\small
	\centering
	\caption{AU partition rule}
	\setlength{\abovecaptionskip}{0pt}
	\setlength{\belowcaptionskip}{2pt}

	\label{tab:AU_region}
\begin{tabular}{lp{15ex}p{22ex}}
	\Xhline{1pt}
	\makecell{AU\\group} & AU NO & RoI NO \\
	\Xhline{1pt}
\makecell[r]{\# $1^{\ast}$ \\ ($\in$\# $2$)} &
\makecell[l]{AU 1 , AU 2 ,\\ AU 5 , AU 7} & \makecell[l]{1, 2, 5, 6, 8, 9, 12, 13,\\ 40, 41, 42, 43} \\
\cmidrule{1-3}
\makecell[r]{\# $2$} &
AU 4 & 1, 2, 3, 4, 5, 6, 8, 9, 12, 13, 40, 41 \\
\cmidrule{1-3}
\makecell[r]{\# $3$ } &
AU 6 & 16, 17, 18, 19, 42, 43 \\
\cmidrule{1-3}
\makecell[c]{\# $4$ \\ ($\in$ \# $3$)} &
AU 9 & 10, 11, 17, 18, 22, 23 \\
\cmidrule{1-3}
\makecell[c]{\# $5$ \\ ($\in$ \# $6$)} &
\makecell[l]{AU 10 , AU 11 ,\\ AU 12 , AU 13 ,\\ AU 14 , AU 15} & \makecell[l]{21, 22, 23, 24, 25, 26,\\ 27, 28, 37} \\
\cmidrule{1-3}
\makecell[c]{\# $6$ \\ ($\in$ \# $5$)} &
\makecell[l]{AU 16 , AU 20 ,\\ AU 25 , AU26 ,\\ AU 27} & \makecell[l]{25, 26, 27, 28, 29, 30,\\ 31, 32, 33, 34, 35, 36, 37} \\
\cmidrule{1-3}
\makecell[c]{\# $7$ \\ ($\in$ \# $6$)} &
AU 17 & \makecell[l]{29, 30, 31, 32, 33, 34, 35,\\ 36} \\
\cmidrule{1-3}
\makecell[c]{\# $8$ \\ ($\in$ \# $5$, \# $6$)} &
\makecell[l]{AU 18 , AU 22 ,\\ AU 23 , AU 24 ,\\ AU 28} &\makecell[l]{26, 27, 29, 30, 31, 32, 37} \\
\Xhline{1pt}

\end{tabular}
\par\vspace{\abovecaptionskip}
{\footnotesize Note: Symbol $\ast$ means the corresponding AU group have symmetrical regions. Symbol $\in$ indicates the ``label fetch".}
\vspace{-0.3cm}
\end{table}
\renewcommand{\arraystretch}{1}

\begin{figure*}[htbp]
	\begin{center}
		\includegraphics[width=1\linewidth]{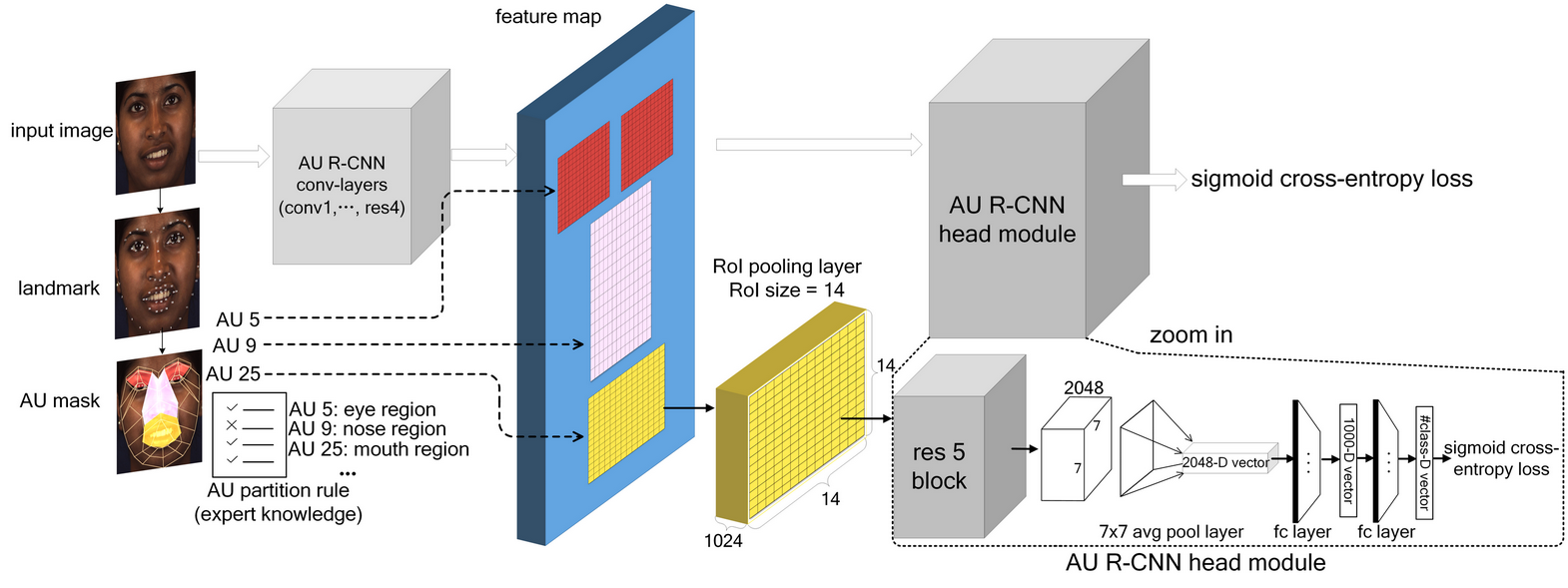}
	\end{center}
	\caption{AU R-CNN using ResNet-101 backbone architecture, where \#class denotes the AU category number we wish to discriminate.}
	\label{fig:AU_R-CNN}
	\vspace{-0.5cm}
\end{figure*}

\subsection{AU R-CNN}
\label{sec:AU_R-CNN}

\begin{figure}[h]
	\setlength{\abovecaptionskip}{0pt}
	\setlength{\abovecaptionskip}{0pt}
	
	\includegraphics[width=0.45\textwidth]{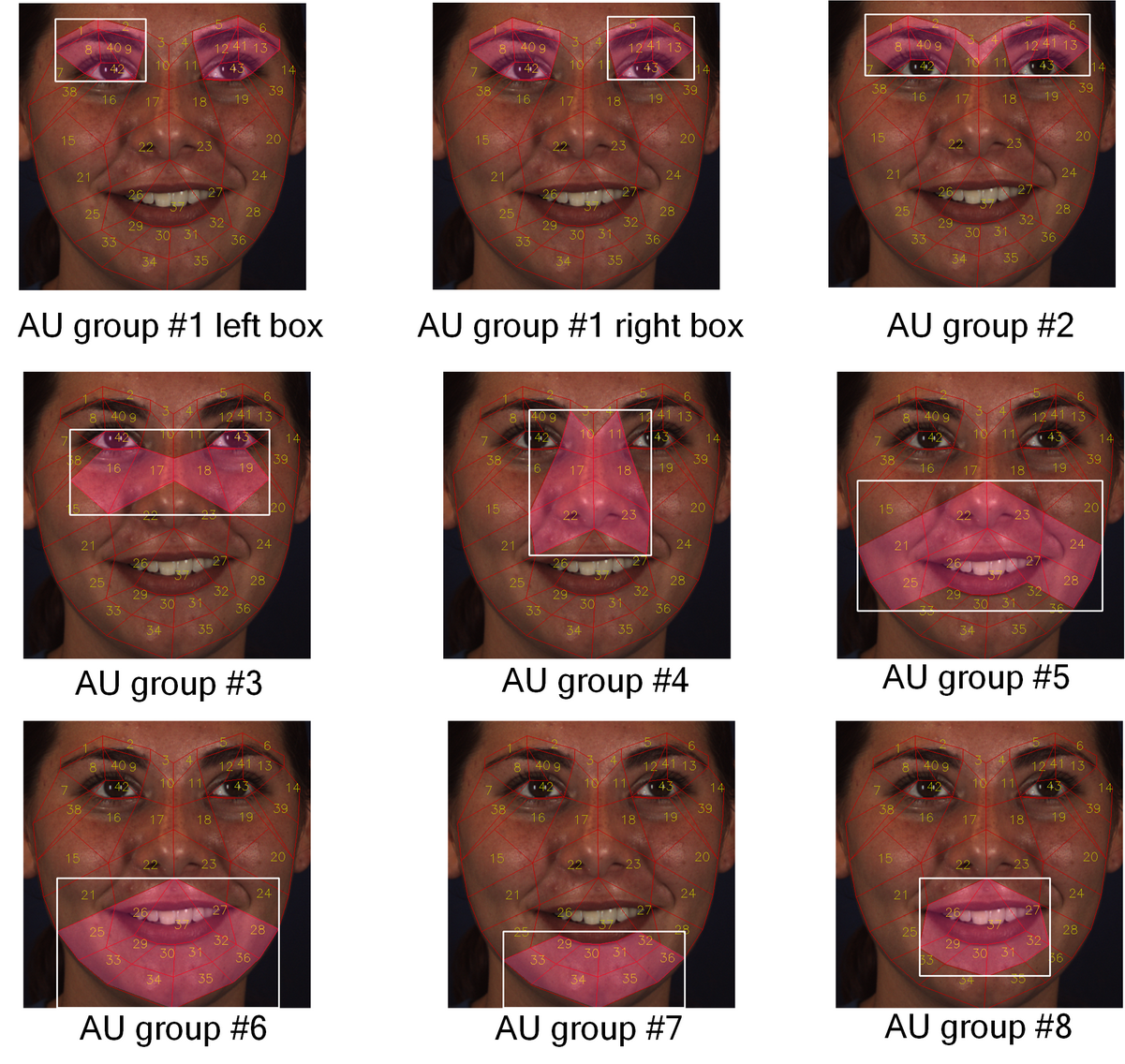}
	
	\caption{AU bounding boxes, which are defined as the minimum bounding box around each AU mask. Since AU group \#1 has two symmetrical regions, the bounding box number is 9.}
	\label{fig:AU_bounding_box}
\end{figure}

AU R-CNN is composed of two modules, namely, feature extraction and head modules. This model can use ResNet \cite{he2016deep} or VGG \cite{Simonyan2014} as its backbone. Here, we use ResNet-101 to illustrate (Fig. \ref{fig:AU_R-CNN}). The feature extraction module comprises conv-layers that produce the feature maps (ResNet-101's conv1, bn1, res2, res3, res4 layers), and the head module includes an RoI pooling layer and the subsequent top layers (res5, avg-pool, and fc layers). After AU masks are obtained, unrelated areas can be excluded. However, each AU mask is an irregular polygon area, which means it cannot be directly fed into the fc layer. Therefore, we introduce the RoI pooling layer originally from Fast R-CNN \cite{girshick2015fast}. The RoI pooling layer is designed to convert the features inside any rectangle RoI (or bounding box) into a small feature map with a fixed spatial extent of $H \times W$. To utilize the RoI pooling layer, each AU mask is converted into a minimum bounding box (named ``AU bounding box") around the mask to input \footnote{AU group \#1 contains two separate symmetrical regions, thus it contains two bounding boxes, which results in total 9 AU bounding boxes, one more than AU group number.}(Fig. \ref{fig:AU_bounding_box}). The RoI pooling layer needs a parameter named ``RoI size", indicates the RoI's height and width after pooling. In our experiment, we set RoI size to $14\times 14$ in ResNet101 backbone and $7\times 7$ in VGG-16 and VGG-19 backbone. 

\begin{figure*}[t]
	\centering
	\includegraphics[width=1\linewidth]{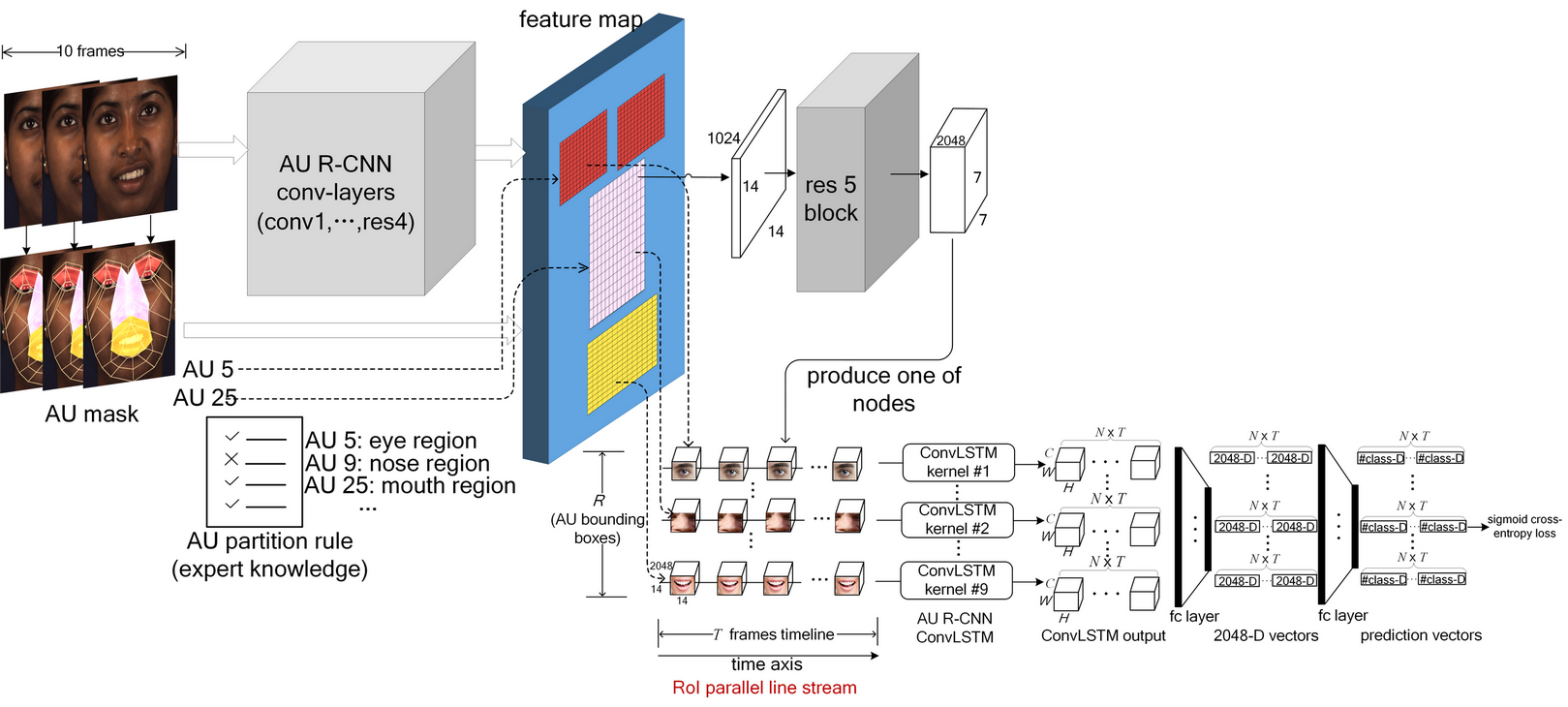}
	\caption{AU R-CNN integrated with ConvLSTM architecture, where $N$ denotes mini-batch size; $T$ denotes the frames to process in each iteration; $R$ denotes AU bounding box number; $C$, $H$, and $W$ denotes the ConvLSTM's output channel number, height and width respectively. \#class denotes the AU category number we wish to discriminate. }
	\label{fig:AR_conv_lstm}
\end{figure*}
Object detection networks, such as Fast R-CNN, aim to identify and localize the object. Benefiting from the design of the AU mask, we have strong confidence in where the AUs should occur; thus, we can concentrate on what the AUs are. Fig. \ref{fig:AU_R-CNN} depicts the AU R-CNN's forward process. In the RoI pooling layer, we input the AU bounding box and feature map (The bounding box coordinates and feature map are usually $16\times$ smaller than the input image resolution). We treat the last fully connected layer's output vector as predicted label probabilities. The total AU category number we wish to discriminate is set as $L$\footnote{$L=22$ in BP4D database and $L=12$ in DISFA database.}; the number of bounding boxes in each image is $R$ \footnote{$R=9$ in BP4D database (Fig. \ref{fig:AU_bounding_box}) and $R=7$ in DISFA database (since DISFA doesn't contain AU group\# 7 and AU group \# 8).}; the ground truth  $\mathbf{y} \in 
\{0,1\}^{R\times L}$, $\mathbf{y}_{i,j}$ indicates the $(i,j)$-th element of $\mathbf{y}$, where $\mathbf{y}_{i,j} = 0$ denotes AU $j$ is inactive in bounding box $i$, and AU $j$ is active if $\mathbf{y}_{i,j} = 1$. The ground truth $\mathbf{y}$ must satisfy the AU partition rule's space constraint: $\mathbf{y}_{i,j} = 0$ if AU $j$ does not belong to bounding box $i$'s corresponding AU group (Fig. \ref{fig:AU_bounding_box} and Table \ref{tab:AU_region}). The RoI-level prediction probability is $\hat{\mathbf{y}} \in \mathbb{R}^{R\times L}$. Given multiple labels inside each RoI (\eg AU 10 and AU 12 often occur together in the mouth area), we adopt the multi-label sigmoid cross-entropy loss function, namely,

\begin{equation}\label{eqn:sigmoid_cross_entropy}
\begin{aligned}
\mathcal{L}(\mathbf{y}, \hat{\mathbf{y}}) =& -\frac{1}{R} \sum_{r=1}^{R}\sum_{l=1}^{L} \{\mathbf{y}_{r,l} \log(\hat{\mathbf{y}}_{r,l}) \}
\end{aligned}
\end{equation}
Unlike ROI-Nets \cite{li2017action} and EAC-Net \cite{li2017eac}, AU R-CNN has considerably fewer parameters due to the sharing of conv-layer in the feature extraction module, which leads to space and time saving. The RoI pooling layer and RoI-level label also help improve classifier performance through the space constraint and supervised information of the RoIs.
 
In the inference stage, the last fc layer's output is converted to a binary integer prediction vector using the threshold of zero (the elements that greater than 0 set to 1, others set to 0). Multiple RoIs' prediction results are merged via a ``bit-wise OR" operator to obtain the image-level label. We report F1 scores of this merged image-level prediction results in Section \ref{sec:Experiment}.

\subsection{Dynamic model extension of AU R-CNN}
\label{sec:method_AR_conv-lstm}
AU R-CNN can use only static RGB images to learn. A natural extension is to use the RoI feature map extracted from AU R-CNN to model the temporal dependency of RoIs across frames. In this extension, we can adopt various dynamic models to observe RoI-level appearance changes (Experiments are shown in Section \ref{sec:static_vs_dynamic}). In this section, we introduce one extension that integrates ConvLSTM \cite{xingjian2015convolutional} into the AU R-CNN architecture.

Fig. \ref{fig:AR_conv_lstm} shows the AU R-CNN integrated with ConvLSTM architecture. In each image, we first extract nine AU group RoI features ($7\times 7\times 2048$) corresponding to nine AU bounding boxes of Fig. \ref{fig:AU_bounding_box} from the last conv-layer. To represent the evolvement of facial local regions, we construct an RoI parallel line stream with nine timelines. The timeline is constructed by skipping four frames per time-step in the video to eliminate the similar frames.  In total, we set 10 time-steps for each iteration. In each timeline, we connect the RoI at the current frame to the corresponding RoI at the adjacent frames, \eg the mouth area has only temporal correlation to the next/previous frame's mouth area. Therefore, each timeline corresponds to an AU bounding box's evolution across time. Nine ConvLSTM kernels are used to process on the nine timelines. The output of each ConvLSTM kernel are fed into two fc layers to produce the prediction probability. More specifically, Let's denote the mini-batch size as $N$. the time-steps as $T$, the channel, height and width of RoI feature as $C$, $H$ and $W$ respectively. The concatenation of ConvLSTM's all time-step's output is a five-dimensional tensor of shape $[N, T, C, H, W]$. We reshape this tensor to a two-dimensional tensor of shape $[N \times T, C \times H \times W]$, the first dimension is treated as the mini-batch of shape $[N\times T]$. This reshaped tensor is input to two fc layers to get a prediction probability vector of shape $[N \times T, Class]$, where $Class$ denotes AU category number.
We adopt the sigmoid cross-entropy loss function to minimize difference between the prediction probability vector and ground truth, which is the same as Eq. \ref{eqn:sigmoid_cross_entropy}. In the inference stage, we use the last frame's prediction result of the 10-frame video clip to evaluate. This model, named ``AR$_{ConvLSTM}$", is trained together with AU R-CNN in an end-to-end form. 

The introduction of the dynamic model extension brings new issues, as shown in our experiments (Section \ref{sec:static_vs_dynamic}), the dynamic model cannot always improve overall performance as expected. We use database statistics and a data visualization technique to identify the effective cases. Various statistics of BP4D and DISFA databases are collected, including the AU duration of each database and the AU group bounding box areas. Li\etal \cite{li2017eac} found that the occurrence of AUs in the database has the influence of static-image-based AU detection classifiers. However, in the ConvLSTM extension model, the average AU activity duration of videos and AR$_{ConvLSTM}$ classification performance are correlated. Fig. \ref{fig:F1_duration_correlation} provides an intuitive figure of such correlation, when the AU duration increases at high peak, the performance of AR$_{ConvLSTM}$ can be always improved. Therefore, in situations such as long-duration activities, AR$_{ConvLSTM}$ can be adopted to improve the performance. Other dynamic models can also be integrated into AU R-CNN, including the two-stream network, TAL-Net, and the general graph CRF model. In Section  \ref{sec:static_vs_dynamic}, we collect the experiment results and analyze various dynamic models in detail. 

\section{Experiments and Results}
\label{sec:Experiment}
\subsection{Settings}
\subsubsection{Dataset description}
We evaluate our method on two datasets, namely, BP4D dataset \cite{zhang2014bp4d} and DISFA dataset \cite{mavadati2013disfa}. For both datasets, we adopt a 3-fold partition to ensure that the subjects are mutually exclusive in the train/test split sets. AUs that present more than $5\%$ base rate are included for evaluation. In total, we select 12 AUs on BP4D and 8 AUs on DISFA to report the experiment results. 

(1) BP4D \cite{zhang2014bp4d} contains $41$ young adults of different races and genders ($23$ females and $18$ males). We use $328$ videos ($41$ participants $\times 8$ videos) captured in total, which result in $\sim \num{140000}$ valid face images. We select positive samples as those with AU intensities equal to or higher than A-level, and the rest are negative samples. We use 3-fold splits exactly the same as \cite{li2017eac,li2017action} partition to ensure that the training and testing subjects are mutually exclusive. The average AU activity duration of all videos in BP4D and the total activity segment count are shown in Table \ref{tab:BP4D_AU_DURATION}. The average AU mask bounding box area is provided in Table \ref{tab:BP4D_AREA}.

(2) DISFA \cite{mavadati2013disfa} contains $27$ subjects. We use $\sim \num{260000}$ valid face images and 54 videos (27 videos captured by left camera and 27 videos captured by right camera).
We also use the 3-fold split partition protocol in the DISFA experiment. The average AU activity duration of all videos in DISFA and the total activity segment count are shown in Table \ref{tab:DISFA_AU_DURATION}. The average AU mask bounding box area is given in Table \ref{tab:DISFA_AREA}.

\subsubsection{Evaluation metric}
Our task is to detect whether the AUs
are active, which is a multi-label binary classification problem. Since our approach focuses on RoI prediction for each bounding box (Fig. \ref{fig:AU_bounding_box}), the RoI-level prediction is a binary vector with $L$ elements, where $L$ denotes the total AU category number we wish to discriminate.  We use the image-level prediction to evaluate, which is obtained by using a ``bit-wise OR" operator for merging an image's RoI-level predictions. After obtaining the image-level prediction, we directly use the database provided image-level ground truth labels to evaluate, which are binary vectors with elements equal 1 for active AUs and equal 0 for inactive AUs. 
The F1 score can be used as an indicator of the performances of the algorithms on each AU and is widely employed in AU detection. In our evaluation, we compute frame-based F1 score \cite{ding2013facial} for 12 AUs in BP4D and 8 AUs in DISFA on image-level prediction. The overall performance of the algorithm is described by the average F1 score(denoted as \textit{Avg.}).

\subsubsection{Compared methods} 
\begin{table}[htp]
	\scriptsize	
	\caption{Compared models details}
	\label{tab:comparative_models}
	\centering
	\tabcolsep=0.15cm
	
	\begin{tabular}{ccccccc}
		\toprule
		\textbf{Model} & \textbf{E2E} & \textbf{ML} & \textbf{RGB} & \textbf{LANDMARK} & \textbf{CONVERGE} & \textbf{VIDEO} \\
		\midrule
		CNN$_{res}$ & $\checkmark$ &  $\checkmark$ & $\checkmark$ & $\times$ & $\checkmark$ & $\times$  \\
		AR$_{vgg16}$ & $\checkmark$ &  $\checkmark$ & $\checkmark$ & $\checkmark$ & $\checkmark$ & $\times$ \\
		AR$_{vgg19}$ & $\checkmark$ &  $\checkmark$ & $\checkmark$ & $\checkmark$ & $\checkmark$ & $\times$ \\
		AR$_{res}$ & $\checkmark$ &  $\checkmark$ & $\checkmark$ & $\checkmark$ & $\checkmark$ & $\times$ \\
		AR$_{mean\_box}$ & $\checkmark$ &  $\checkmark$ & $\checkmark$ & $\times$ & $\checkmark$ & $\times$ \\
		AR$_{FPN}$ & $\checkmark$ &  $\checkmark$ & $\checkmark$ & $\checkmark$ & $\checkmark$ & $\times$ \\
		AR$_{ConvLSTM}$ & $\checkmark$ & $\checkmark$ & $\checkmark$ & $\checkmark$ & $\checkmark$ & $\checkmark$ \\
		AR$_{2stream}$ & $\checkmark$ & $\checkmark$ & $\times$ & $\checkmark$ & $\checkmark$ & $\checkmark$ \\
		AR$_{CRF}$ & $\times$ & $\times$ & $\checkmark$ & $\checkmark$ & $\checkmark$ & $\checkmark$ \\
		AR$_{TAL}$ & $\times$ & $\checkmark$ & $\times$ & $\checkmark$ & $\times$ & $\checkmark$ \\
		\bottomrule
	\end{tabular}
	
	{\footnotesize * \textbf{E2E}: end-to-end trainable, \textbf{ML}: multi-label learning, \textbf{RGB}: only use RGB information, not incorporate optical flow, \textbf{LANDMARK}: use landmark point, \textbf{CONVERGE}: the model converged in training, \textbf{VIDEO}: need video context.}
\end{table}

We collect the F1 scores of the most popular state-of-the-art approaches that used the same 3-fold protocol in Table \ref{tab:BP4D_F1} and Table \ref{tab:DISFA_F1} to compare our approaches with other methods. These techniques include a linear support vector machine (LSVM), active patch learning (APL \cite{zhong2015learning}), JPML \cite{Zhao2016}, a confidence-preserving machine (CPM \cite{zeng2015confidence}), a block-based region learning CNN (DRML \cite{Zhao2016b}), an enhancing and
cropping nets (EAC-net \cite{li2017eac}), an ROI adaption net (ROI-Nets \cite{li2017action}), and LSTM fused with a simple CNN (CNN+LSTM \cite{Chu2017Learning}), an optimized filter size CNN (OFS-CNN \cite{han2017optimizing}). We also conduct complete control experiments of AU R-CNN in Table \ref{tab:BP4D_control_expr} and Table \ref{tab:DISFA_control_expr}, including ResNet-101 based traditional CNN that classifies the entire face images (CNN$_{res}$), ResNet-101 based AU R-CNN (AR$_{res}$), VGG-16 based AU R-CNN (AR$_{vgg16}$), VGG-19 based AU R-CNN (AR$_{vgg19}$), mean bounding boxes version AU R-CNN (AR$_{mean\_box}$), AU R-CNN incorporate with Feature Pyramid Network \cite{lin2017feature}(AR$_{FPN}$), AU R-CNN integrated with ConvLSTM \cite{xingjian2015convolutional} (AR$_{ConvLSTM}$),
AU R-CNN with optical flow and RGB feature fusion two-stream network architecture \cite{feichtenhofer2016convolutional}(AR$_{2stream}$), general graph CRF with features extracted by AU R-CNN(AR$_CRF$), and AU R-CNN with a temporal action localization in video network, TAL-Net \cite{chao2018rethinking}(AR$_{TAL}$). We use ResNet-101 based CNN(CNN$_{res}$) as our baseline model. The details of the compared models are summarized in Table \ref{tab:comparative_models}.

\subsubsection{Implementation details} 
We resize the face images to $512 \times 512$ after cropping the face areas. Each image and bounding boxes are horizontally mirrored randomly before being sent to AU R-CNN for data augmentation. We subtract the mean pixel value from all the images in the dataset before sending to AU R-CNN. We use dlib \cite{king2009dlib} to landmark faces, and the landmark operator is consequently time consuming. We cache the mask in the memcached database to accelerate speed in later epochs. The VGG and ResNet-101 backbones of AU R-CNN use pre-trained ImageNet ILSVRC dataset \cite{deng2009imagenet} weights to initialize. AU R-CNN is initialized with a learning rate of 0.001 and further reduced by a factor of 0.1 after every 10 epochs. In all experiments, we select momentum stochastic gradient descent to train AU R-CNN for 25 epochs and set momentum to 0.9 and weight decay to 0.0005. The mini-batch size is set to 5.
\begin{table*}[htp]

	\scriptsize
	\setlength{\abovecaptionskip}{0pt} 
	
	\caption{F1 score result comparison with state-of-the-art methods on \textbf{BP4D} dataset. Bracketed bold numbers indicate the best score; bold numbers indicate the second best.}
	\label{tab:BP4D_F1}
	\centering
	\tabcolsep=0.1cm
	\begin{tabular}{c*{9}{c}ccc}
		\toprule
		AU & LSVM & JPML \cite{Zhao2016} & DRML \cite{Zhao2016b} & CPM \cite{zeng2015confidence} & CNN+LSTM \cite{Chu2017Learning} & EAC-Net \cite{li2017eac} & OFS-CNN \cite{han2017optimizing} & ROI-Nets \cite{li2017action} & FERA \cite{jaiswal2016deep} & AR$_{vgg16}$ & AR$_{vgg19}$ & AR$_{res}$ \\
		\midrule
1 & 23.2 & 32.6 & 36.4 & 43.4 & 31.4 & 39 & 41.6 & 36.2 & 28 & \textbf{47.5} & 44.8 & [\textbf{50.2}] \\
2 & 22.8 & 25.6 & 41.8 & 40.7 & 31.1 & 35.2 & 30.5 & 31.6 & 28 & 40.5 & \textbf{43.5} & [\textbf{43.7}] \\
4 & 23.1 & 37.4 & 43 & 43.4 & [\textbf{71.4}] & 48.6 & 39.1 & 43.4 & 34 & 55.1 & 52.2 & \textbf{57} \\
6 & 27.2 & 42.3 & 55 & 59.2 & 63.3 & 76.1 & 74.5 & \textbf{77.1} & 70 & 73.8 & 75.7 & [\textbf{78.5}] \\
7 & 47.1 & 50.5 & 67 & 61.3 & 77.1 & 72.9 & 62.8 & 73.7 & \textbf{78} & 76.6 & 75.2 & [\textbf{78.5}] \\
10 & 77.2 & 72.2 & 66.3 & 62.1 & 45 & 81.9 & 74.3 & [\textbf{85}] & 81 & 82 & \textbf{82.7} & 82.6 \\
12 & 63.7 & 74.1 & 65.8 & 68.5 & 82.6 & 86.2 & 81.2 & [\textbf{87}] & 78 & 85.2 & 85.9 & [\textbf{87}] \\
14 & 64.3 & 65.7 & 54.1 & 52.5 & \textbf{72.9} & 58.8 & 55.5 & 62.6 & [\textbf{75}] & 64.9 & 63.4 & 67.7 \\
15 & 18.4 & 38.1 & 36.7 & 34 & 34 & 37.5 & 32.6 & 45.7 & 20 & \textbf{48.8} & 45.3 & [\textbf{49.1}] \\
17 & 33 & 40 & 48 & 54.3 & 53.9 & 59.1 & 56.8 & 58 & 36 & \textbf{60.6} & 60 & [\textbf{62.4}] \\
23 & 19.4 & 30.4 & 31.7 & 39.5 & 38.6 & 35.9 & 41.3 & 38.3 & 41 & 43.9 & \textbf{46.1} & [\textbf{50.4}] \\
24 & 20.7 & 42.3 & 30 & 37.8 & 37 & 35.8 & - & 37.4 & - & [\textbf{49.3}] & 48.3 & [\textbf{49.3}] \\
\midrule
Avg & 35.3 & 45.9 & 48.3 & 50 & 53.2 & 55.9 & 53.7 & 56.4 & 51.7 & \textbf{60.7} & 60.3 & [\textbf{63}] \\
\bottomrule
	\end{tabular}
\end{table*}

 \begin{table*}[!htbp]
	\scriptsize
	\centering
	
	\setlength{\abovecaptionskip}{0pt} 
	
	\caption{\textbf{Control experiments for BP4D}. Results are reported using F1 score on 3-fold protocol.}
	\label{tab:BP4D_control_expr}
	
	\centering
	\tabcolsep=0.05cm
	\begin{tabular}{c*{10}{c}}
		
		\toprule
		AU & CNN$_{res}$ & AR$_{vgg16}$ & AR$_{vgg19}$ & AR$_{res}$ & AR$_{mean\_box}$ & AR$_{FPN}$ & AR$_{ConvLSTM}$ & AR$_{2stream}$ & AR$_{CRF}$ & AR$_{TAL}$ \\
		\midrule
		1 & 45.8 & 47.5 & 44.8 & [\textbf{50.2}] & 45.8 & 46.4 & 48 & 46.6& \textbf{50.1}& 41.3 \\
		2 & 43.2 & 40.5 & \textbf{43.5} & [\textbf{43.7}] & 41.1 & 40.7 & 43.2 & 42.1 &35 & 37.4 \\
		4 & 54.3 & 55.1 & 52.2 & [\textbf{57}] & [\textbf{57}] & 47.5 & 53.1 & 52.4 &45.2 & 44.5 \\
		6 & \textbf{77.4} & 73.8 & 75.7 & [\textbf{78.5}] & 75.1 & 76.4 & 76.9 & 75.4 &71.4 & 64.4 \\
		7 & 77.9 & 76.6 & 75.2 & [\textbf{78.5}] & 77.7 & 76.9 & \textbf{78.4} & 77.3 & 77.7 & 73.6 \\
		10 & 81.8 & 82 & \textbf{82.7} & 82.6 & 82.2 & 81.3 & [\textbf{82.8}] & 82.1 & 82.1 & 76.2 \\
		12 & 85.8 & 85.2 & 85.9 & 87 & 86.5 & 85.4 & [\textbf{87.9}] & \textbf{87.1} &86.9 & 80 \\
		14 & 60.8 & 64.9 & 63.4 & [\textbf{67.7}] & 62 & 63.5 & [\textbf{67.7}] & 62.7 &67.2 & 64.9 \\
		15 & [\textbf{50}] & 48.8 & 45.3 & 49.1 & 48 & 44.9 & 45.6 & \textbf{49.6}& 47.6& 45.7 \\
		17 & 58.3 & 60.6 & 60 & 62.4 & 61.5 & 57.9 & [\textbf{63.4}] & \textbf{63.2} & 58.7& 53.3 \\
		23 & 47.6 & 43.9 & 46.1 & [\textbf{50.4}] & 48.7 & 42.3 & 47.9 & \textbf{49.9} &36.8& 39.1 \\
		24 & 48.4 & 49.3 & 48.3 & 49.3 & 53.2 & 46.6 & \textbf{56.4} & [\textbf{57.6}] &51.6 & 49.5 \\
		\midrule
		Avg & 60.9 & 60.7 & 60.3 & [\textbf{63}] & 61.6 & 59.2 & \textbf{62.6} & 62.2 &59.2& 55.8 \\
		\bottomrule
	\end{tabular}
	\vspace{0.0cm}
\end{table*}

\subsection{Conventional CNN versus AU R-CNN}

\begin{figure}
	\setlength{\abovecaptionskip}{0pt}
	\setlength{\belowcaptionskip}{0pt}
	\begin{adjustbox}{max width=0.5\textwidth,center}
		\subfigure[AU 1, 4, 7, 10]{
			\includegraphics[width=0.11\textwidth]{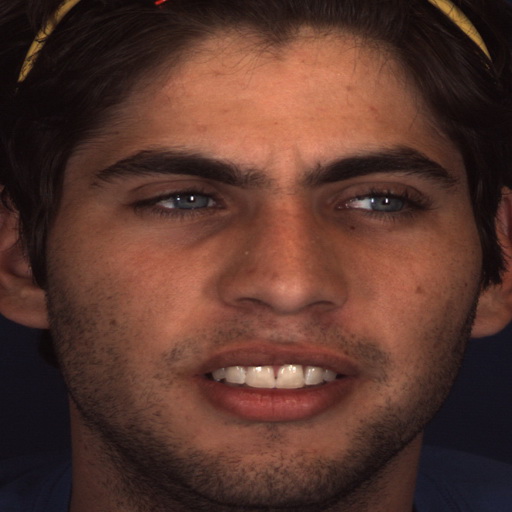}
		}
		\subfigure[AU 6, 7, 10, 12]{
			\includegraphics[width=0.11\textwidth]{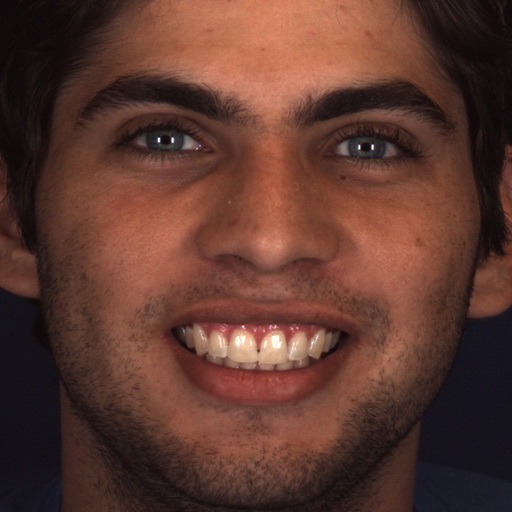}
		}
		\subfigure[AU 2, 14, 17]{
			\includegraphics[width=0.11\textwidth]{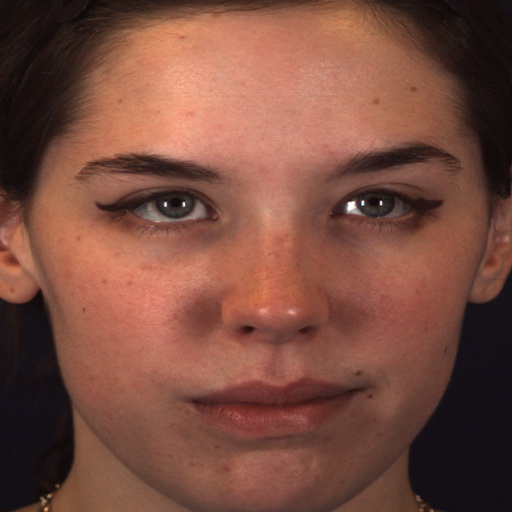}
		}
		\subfigure[AU 6, 7, 12]{
			\includegraphics[width=0.11\textwidth]{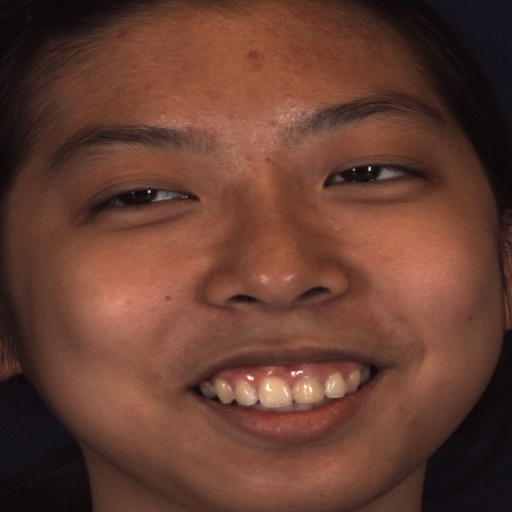}
		}
		\subfigure[AU 7, 14, 17]{
			\includegraphics[width=0.11\textwidth]{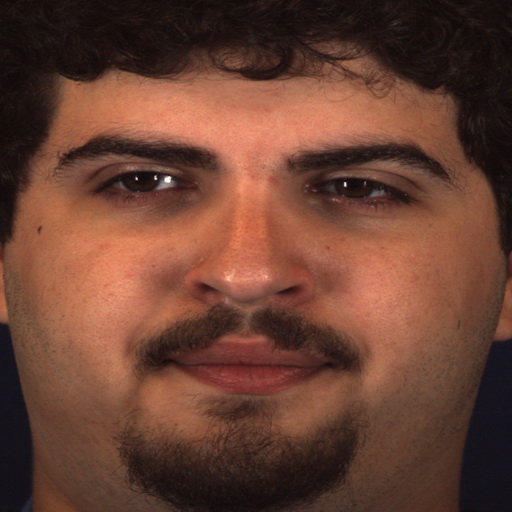}
		}
	\end{adjustbox}
	\caption{Example figures of detection result.}
	\label{fig:detection_result}
\end{figure}

AU R-CNN is proposed for adaptive regional learning in Section \ref{sec:AU_R-CNN}. Thus, our first experiment aims to determine whether it can perform better than the baseline conventional CNN, which uses entire face images to learn. We suppose that by learning the adaptive RoIs separately, recognition capability can be improved. We train CNN$_{res}$ and AR$_{res}$ on the BP4D and DISFA datasets using the same ResNet-101 backbone for comparison. Twelve AUs in BP4D and eight AUs in DISFA are used; therefore, AR$_{res}$ and CNN$_{res}$ use the sigmoid cross-entropy loss function, as shown in Eq. \ref{eqn:sigmoid_cross_entropy}. Both models are based on static images. During each iteration, we randomly select five images to comprise one mini-batch to train and initialize the learning rate to 0.001.

\begin{table}
	\scriptsize
	\centering
	
	\setlength{\abovecaptionskip}{0pt} 
	
	\caption{F1 score of varying resolutions comparison result on \textbf{BP4D} dataset. The \textbf{bold} highlights the best performance in each resolution experiment.}
	\label{tab:BP4D_resolution}
	\centering
	\tabcolsep=0.05cm
	\begin{tabular}{ccc @{\hspace{5\tabcolsep}} cc @{\hspace{5\tabcolsep}} cc @{\hspace{5\tabcolsep}} cc}
		
		\toprule
		resolution & \multicolumn{2}{c}{$256 \times 256$} & \multicolumn{2}{c}{$416 \times 416$} & \multicolumn{2}{c}{$512 \times 512$} & \multicolumn{2}{c}{$608 \times 608$} \\
		\midrule
		AU & CNN$_{res}$ & AR$_{res}$ & CNN$_{res}$ & AR$_{res}$ & CNN$_{res}$ & AR$_{res}$ & CNN$_{res}$ & AR$_{res}$ \\
		\midrule
		1 & 45.6 & \textbf{50.1} & 47.4 & \textbf{49.3} & 45.8 & \textbf{50.2} & 44.3 & \textbf{47.5} \\
		2 & 43.6 & \textbf{46.5} & 38.3 & \textbf{42.1} & 43.2 & \textbf{43.7} & \textbf{40.1} & 39.2 \\
		4 & 52.2 & \textbf{54.6} & \textbf{53.3} & 50.0 & 54.3 & \textbf{57.0} & 49.5 & \textbf{53.5} \\ 
		6 & 74.9 & \textbf{77.7} & \textbf{75.7} & 75.2 & 77.4 & \textbf{78.5} & 76.3 & \textbf{76.9} \\
		7 & 76.3 & \textbf{78.3} & 75.7 & \textbf{78.7} & 77.9 & \textbf{78.5} & 76.4 & \textbf{78.6} \\
		10 & \textbf{82.5} & 81.7 & \textbf{82.4} & 82.3 & 81.8 & \textbf{82.6} & 81.5 & \textbf{82.7} \\
		12 & 86.5 & \textbf{87.5} & \textbf{87.2} & 86.5 & 85.8 & \textbf{87.0} & \textbf{87.5} & 85.5 \\
		14 & 55.4 & \textbf{62.1} & 59.5 & \textbf{61.9} & 60.8 & \textbf{67.7} & 59.5 & \textbf{62.0} \\
		15 & 48.0 & \textbf{51.2} & 44.1 & \textbf{49.2} & \textbf{50.0} & 49.1 & 44.9 & \textbf{49.6} \\
		17 & 59.9 & \textbf{61.8} & 57.5 & \textbf{61.4} & 58.3 & \textbf{62.4} & 57.4 & \textbf{61.3} \\
		23 & 44.7 & \textbf{46.2} & 41.2 & \textbf{44.9} & 47.6 & \textbf{50.4} & \textbf{45.6} & 45.1 \\
		24 & 46.9 & \textbf{52.3} & 44.5 & \textbf{47.7} & 48.4 & \textbf{49.3} & 48.2 & \textbf{51.1} \\
		\midrule
		Avg & 59.7 & \textbf{62.5} & 58.9 & \textbf{60.8} & 60.9 & \textbf{63.0} & 59.3 & \textbf{61.1} \\
		\bottomrule
	\end{tabular}
	\vspace{-0.5cm}
\end{table}

Fig. \ref{fig:detection_result} demonstrates the example detection results of our approach.
Table \ref{tab:BP4D_control_expr} and Table \ref{tab:DISFA_control_expr} show the BP4D and DISFA results, in which the margin is larger in DISFA (3.69\%) than in BP4D (2.1\%). These results can be attributed to the relatively lower resolution images in DISFA, which cause AR$_{res}$ to benefit more.
We also show that AU R-CNN performs efficiently with varying image resolutions. Experiments have been conducted to compare the proposed AU R-CNN and baseline CNN with the same ResNet-101 backbone on the BP4D database with different resolutions of the input image. Table \ref{tab:BP4D_resolution} shows the result, and the resolutions of images are set to $256\times 256$, $416 \times 416$, $512 \times 512$, and $608 \times 608$. Most AU results prefer AU R-CNN model by observing subtle cues of facial appearance changes. In $256 \times 256$, although the resolution is nearly half of that in $512 \times 512$, the performance is close to that in $512 \times 512$. This similarity leads to efficient detection when using $256 \times 256$. But in the highest resolution $608 \times 608$, the F1 score is lower than that of $512 \times 512$, we believe this performance drop can be attribute to two possible reasons. (1)
As pointed out by Han \etal \cite{han2017optimizing}, when the image resolution increases to $608 \times 608$, the receptive field covers a smaller actual area of the entire face when using the same convolution filter size. The smaller receptive field deteriorates the vision. (2) Larger images produce larger feature maps before RoI pooling layer in AR$_{res}$, or larger feature maps before the final avg pooling layer in CNN$_{res}$. The increase of feature map size also increases each pooling grid cell's covered size dramatically in RoI pooling/avg pooling layer, which has negative impact on high level features. Regardless of the overall improvement of AU R-CNN across the 12 AUs. In AU 10 and AU 12, CNN and AU R-CNN obtain similar results. One explanation is that AU 10 and AU 12 have relatively sufficient training samples compared with other AUs.

In the DISFA dataset (Table \ref{tab:DISFA_control_expr}, Table \ref{tab:DISFA_F1}), AR$_{res}$ outperforms CNN$_{res}$ in six out of eight AUs. The two remaining AUs are AU 12 and AU 25. As shown in Table \ref{tab:DISFA_AREA}, AU 12 and AU 25 have the largest area proportions (29.8 \% and 26.6 \%) on the face images. In BP4D and DISFA, AU 1 (inner brow raiser) has a significant improvement in AR$_{res}$ because of the relatively small area on the face.
 \begin{table}
	\scriptsize
	\centering
	
	\setlength{\abovecaptionskip}{0pt} 
	
	\caption{F1 score result comparison with state-of-the-art methods on \textbf{DISFA} dataset. Bracketed bold numbers indicate the best score; bold numbers indicate the second best.}
	\label{tab:DISFA_F1}
	
	\centering
	\tabcolsep=0.05cm
	\begin{tabular}{c*{6}{c}cc}
		
		\toprule
		AU & LSVM & APL \cite{zhong2015learning} & DRML \cite{Zhao2016b} & ROI-Nets \cite{li2017action} & CNN$_{res}$ & AR$_{vgg16}$ & AR$_{vgg19}$ & AR$_{res}$ \\
		\midrule
	1 & 10.8 & 11.4 & 17.3 & [\textbf{41.5}] & 26.3 & 24.9 & 26.9 & \textbf{32.1} \\
	2 & 10 & 12 & 17.7 & [\textbf{26.4}] & 23.4 & 23.5 & 21 & \textbf{25.9} \\4 & 21.8 & 30.1 & 37.4 & [\textbf{66.4}] & 51.2 & 55.5 & 59.6 & \textbf{59.8} \\
	6 & 15.7 & 12.4 & 29 & 50.7 & 48.1 & 51 & [\textbf{56.5}] & \textbf{55.3} \\9 & 11.5 & 10.1 & 10.7 & 8.5 & 29.9 & \textbf{41.8} & [\textbf{46}] & 39.8 \\
	12 & \textbf{70.4} & 65.9 & 37.7 & [\textbf{89.3}] & 69.4 & 68 & 67.7 & 67.7 \\
	25 & 12 & 21.4 & 38.5 & [\textbf{88.9}] & \textbf{80.1} & 74.9 & 79.8 & 77.4 \\26 & 22.1 & 26.9 & 20.1 & 15.6 & \textbf{52.4} & 49.4 & 47.6 & [\textbf{52.6}] \\
	\midrule
	Avg & 21.8 & 23.8 & 26.7 & 48.5 & 47.6 & 48.6 & \textbf{50.7} & [\textbf{51.3}] \\
		\bottomrule
	\end{tabular}
	\vspace{-0.0cm}
\end{table}

\begin{table*}
	\scriptsize
	\centering
	
	\setlength{\abovecaptionskip}{0pt} 
	
	\caption{\textbf{Control experiments for DISFA}. Results are reported using F1 score on 3-fold protocol.}
	\label{tab:DISFA_control_expr}

	\centering
	\tabcolsep=0.05cm
	\begin{tabular}{c*{9}{c}}
		
		\toprule
		AU & CNN$_{res}$ & AR$_{vgg16}$ & AR$_{vgg19}$ & AR$_{res}$ & AR$_{mean\_box}$ & AR$_{FPN}$ & AR$_{ConvLSTM}$ & AR$_{2stream}$ & AR$_{CRF}$ \\
		\midrule
		1 & 26.3 & 24.9 & 26.9 & 32.1 & 31.3 & [\textbf{39.9}] & 26.9 & \textbf{34.3} & 24.1  \\
		2 & 23.4 & 23.5 & 21 & 25.9 & \textbf{28.3} & [\textbf{33.3}] & 24.4 & 27.4 & 26.5  \\
		4 & 51.2 & 55.5 & \textbf{59.6} & [\textbf{59.8}] & 59.3 & 59.3 & 58.6 & 59.4 & 51.7 \\
		6 & 48.1 & 51 & 56.5 & 55.3 & 55.4 & 49.3 & 49.7 & [\textbf{59.8}] & \textbf{57.8} \\
		9 & 29.9 & 41.8 & [\textbf{46}] & 39.8 & 38.4 & 32.5 & 34.2 & \textbf{42.1} & 33 \\
		12 & \textbf{69.4} & 68 & 67.7 & 67.7 & 67.7 & 65.5 & [\textbf{71.3}] & 65 & 65.5\\
		25 & \textbf{80.1} & 74.9 & 79.8 & 77.4 & 77.2 & 72.6 & [\textbf{83.4}] & 77.4 & 71 \\
		26 & 52.4 & 49.4 & 47.6 & 52.6 & \textbf{52.8} & 47.9 & 51.4 & 50.1 & [\textbf{53.5}] \\
		\midrule
		Avg & 47.6 & 48.6 & 50.7 & \textbf{51.3} & \textbf{51.3} & 50 & 50 & [\textbf{51.9}] & 47.9 \\
		\bottomrule
	\end{tabular}
	\vspace{0.0cm}
\end{table*}

\subsection{ROI-Nets versus AU R-CNN}

\begin{table}[htp]
	\scriptsize	
	\setlength{\abovecaptionskip}{0pt}
	\setlength{\abovecaptionskip}{0pt}
	\caption{Average bounding box area in \textbf{BP4D}}
	\label{tab:BP4D_AREA}
	\centering
	\tabcolsep=0.15cm
	\begin{tabular}{c*{6}{p{6.7ex}}}
		\toprule
		AU group & \# 1 & \# 2 & \# 3 & \# 5 & \# 7 & \# 8\\
		\midrule
		AU index & 1,2,7 & 4 & 6 & 10,12, 14,15 & 17 & 23,24 \\
		\midrule
		Avg box area (pixels) & 17785 & 46101 & 54832 & 103875 & 42388 & 38470 \\
		Area proportion & 6.8\% & 17.6\% & 20.9\% & 39.6 \% & 16.2 \% & 14.7 \% \\
		\bottomrule
	\end{tabular}
	\vspace{1cm}
\end{table}

\begin{table}[htp]
	\scriptsize	
	\setlength{\abovecaptionskip}{0pt}
	\setlength{\abovecaptionskip}{0pt}
	\caption{Average bounding box area in \textbf{DISFA}}
	\label{tab:DISFA_AREA}
	\centering
	\tabcolsep=0.15cm
	\begin{tabular}{c*{6}{p{6.7ex}}}
		\toprule
		AU group & \# 1 & \# 2 & \# 3 & \# 4 & \# 5 & \# 6 \\
		\midrule
		AU index & 1,2 & 4 & 6 & 9 & 12 & 25,26 \\
		\midrule
		Avg box area (pixels) & 17545 & 45046 & 46317 & 48393 & 78131 & 69624\\
		Area proportion & 6.7\% & 17\% & 17.7\% & 18.5 \% & 29.8 \% & 26.6 \% \\
		\bottomrule
	\end{tabular}
	\vspace{1cm}
\end{table}

\begin{figure}[htbp]
	\centering
	\setlength{\abovecaptionskip}{0pt}
	\setlength{\abovecaptionskip}{0pt}
	\includegraphics[width=0.15\textwidth]{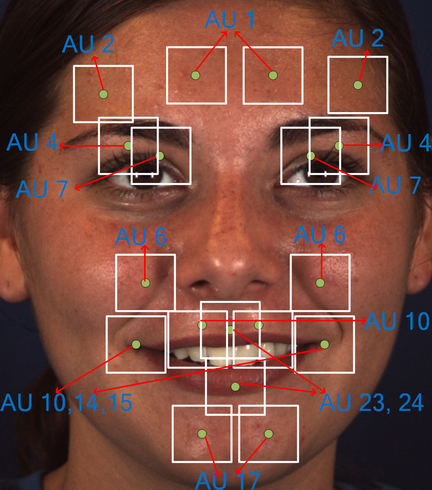}
	\caption{The AU centers of ROI-Nets, each AU center location is an offset of landmark point, and the $3\times 3$ bounding boxes centered at AU centers from top layer's feature map are cropped.}
	\label{fig:AU_center}
\end{figure}

Our proposed AU R-CNN in Section \ref{sec:AU_R-CNN} is designed to recognize local regional AUs in static images under AU mask. Previous state-of-the-art static image AU detection approach ROI-Nets \cite{li2017action} also focuses on regional learning. It attempts to learn regional features by using individual conv-layers over regions centered on AU center (Fig. \ref{fig:AU_center}). The two models are based on static images, whereas our AU R-CNN uses the shared conv-layer in feature extraction module and RoI-level supervised information. This choice saves space and time, and provides accurate guidance. Instead of using the concept of the AU center area, we introduce the AU mask. We believe that AU mask can preserve more context information than cropping the bounding box from AU center. ROI-Nets adopts VGG-19 as backbone. For fair comparison, we adopt VGG-19 based AU R-CNN (denoted as AR$_{vgg19}$) to compare. AR$_{vgg19}$ outperforms ROI-Nets in 8 out of 12 AUs in BP4D (Table \ref{tab:BP4D_F1}).  

The interesting part lies in AU 23 (lip pressor) and AU 24 (lip tighter), in which AR$_{vgg19}$ significantly outperforms ROI-Nets by 7.8\% and 10.9\%, respectively. This superiority is because the lip area is a relatively small area on face; AU R-CNN uses AU mask and RoI-level label so that it can concentrate on this area. This fact can be verified from Table \ref{tab:BP4D_AREA} that the AU 23 and AU 24 bounding box only occupies 14.7\% area of the face image. 
Other typical cases are AU 1, AU 2, and AU 4, which are located in the areas around eyebrows and eyes; AR$_{vgg19}$ outperforms ROI-Nets by 8.5\%, 11.9\%, and 8.8\%, respectively. In AU 6 (cheek raiser, see Fig. \ref{fig:mask_3}), AU 10, AU 12, AU 14, and AU 15 results, ROI-Nets and AU R-CNN achieve close results. These areas occupy relatively large proportions in the image (Table \ref{tab:BP4D_AREA}), and ROI-Nets focuses on the central large area of the image. The experiment in DISFA dataset (Table \ref{tab:DISFA_F1}) demonstrates the similar result. The above comparisons prove that, AU R-CNN better expresses the classification information of local regions than ROI-Nets. We also found that the ResNet-based AU R-CNN (AR$_{res}$) outperforms AR$_{vgg19}$ in the BP4D and DISFA datasets, and achieves the best performance over all static-image-based approaches. For better representation of AU features, we conduct our remaining experiments on the basis of AR$_{res}$ features.

\begin{table}[htp]
	\scriptsize	
	\setlength{\abovecaptionskip}{0pt}
	\setlength{\abovecaptionskip}{0pt}
	\caption{Inference time(ms) of VGG-19 on $512\times 512$ images}
	\label{tab:running_time}
	\centering
	\tabcolsep=0.15cm
	\begin{tabular}{ccc}
		\toprule
		Ours & ROI-Nets \cite{li2017action} & LCN \cite{taigman2014deepface} \\
		\midrule
		$27.4 \pm 0.0005$ & $67.7 \pm 0.0004$ & $34.7 \pm 0.008$ \\
		\bottomrule
	\end{tabular}
\end{table}
We further evaluate the inference time of our approach, LCN (CNN with locally connected layer \cite{taigman2014deepface}) and ROI-Nets on a Nvidia Geforce GTX 1080Ti GPU. We run each network for 20 trails over \num{1000} iterations with the mini-batch size sets to 1; then we evaluate the running time for each iteration, and finally compute the mean and standard deviation over the 20 trials. The inference time is showed in Table \ref{tab:running_time}, we can see our approach benefits from the RoI pooling layer's parallel computing over multiple bounding boxes, its inference time is lower than LCN and ROI-Nets. The RoI-Nets adopt 20 individual conv-layers for 20 bounding boxes, thus it results worst performance.

\subsubsection{AU R-CNN + Mean Box}
\label{sec:mean_box}
\begin{table}[htp]
	\scriptsize	
	\setlength{\abovecaptionskip}{0pt} 
	\caption{Mean box coordinates of $512\times 512$ images in \textbf{BP4D}.}
	\label{tab:BP4D_mean_box}
	\centering
	\tabcolsep=0.05cm
	\begin{tabular}{lll}
		\toprule
		
		AU group & AU index & Mean boxes coordinates ($y_{min}$, $x_{min}$, $y_{max}$, $x_{max}$ format) \\
		\midrule
		\# 1 & 1,2,7 & (30.4, 58.1, 140.3, 222.5), (30.1, 297.2, 140.9, 456.5)  \\
		\# 2 & 4 & (23.9, 57.8, 139, 455.9) \\
		\# 3 & 6 & (109.4, 79.8, 264.5, 431.8) \\
		\# 5 & 10,12,14,15 & (198.9, 35.2, 437.0, 472.6) \\
		\# 7 & 17 & (378.7, 94.5, 510.9, 416.6) \\
		\# 8 & 23,24 & (282.7,145.5,455.0,368.3) \\
		\bottomrule
	\end{tabular}
	\vspace{-0.0cm}
\end{table}

\begin{table}[htp]
	\scriptsize	
	\setlength{\abovecaptionskip}{0pt} 
	\caption{Mean box coordinates of $512\times 512$ images in \textbf{DISFA}.}
	\label{tab:DISFA_mean_box}
	\centering
	\tabcolsep=0.05cm
	\begin{tabular}{lll}
		\toprule
		
		AU group & AU index & Mean boxes coordinates ($y_{min}$, $x_{min}$, $y_{max}$, $x_{max}$ format) \\
		\midrule
		\# 1 & 1,2 & (55.5, 71.3, 168.6, 220.0), (53.5, 277.6, 167.6, 431.4)  \\
		\# 2 & 4 & (48.5, 58.7, 165.1, 444.0) \\
		\# 3 & 6 & (141.4, 86.7, 281.5, 418.9) \\
		\# 4 & 9 & (107.8, 152.2, 348.8, 352.8) \\
		\# 5 & 12 & (236.9, 53.5, 433.3, 454.4) \\
		\# 6 & 25,26 & (316.4, 73.8, 511.0, 433.4) \\
		\bottomrule
	\end{tabular}
\end{table}
The computation of each image's precise landmark point location is time consuming. We believe it is enough to use the ``mean" AU bounding box coordinates to represent all images' bounding boxes. In this section, we collect the average coordinates of all images of nine AU group bounding boxes in each database to form a unified ``mean box" across all images (Table \ref{tab:BP4D_mean_box} and Table \ref{tab:DISFA_mean_box}). We use this ``mean box" coordinates to replace the real bounding box coordinates calculated from the landmark in each image to evaluate. The experiment results are shown in Table \ref{tab:BP4D_control_expr} and Table \ref{tab:DISFA_control_expr}, denoted as AR$_{mean\_box}$. Although most images of BP4D and DISFA dataset are the frontal face, the deviation of mean bounding box coordinates from real box location exists. However, the F1 score is remarkably close to AR$_{res}$, because the RoI pooling layer in AU R-CNN performs a coarse spatial quantization. This performance similarity demonstrates that AU R-CNN is robust to small landmark location error, and the computation consumption of each image's landmark can be saved via using ``mean box".

\subsection{AU R-CNN + Feature Pyramid Network}
In the previous sections, we use the single scale ($16\times$ smaller scale) RoI feature to detect. Feature Pyramid Network (FPN) \cite{lin2017feature} is a popular architecture for leveraging a CNN's pyramidal features in the object detection field, which has semantics from low to high levels. In this experiment, FPN is integrated into AU R-CNN's backbone as feature extractor that extracts RoI features from the feature pyramid. The assignment of an RoI of width $w$ and height $h$ to the level $k$ of FPN is as follows \cite{lin2017feature}:

\begin{equation}
\begin{aligned}
\label{eqn:FPN_level}
k = \lceil k_0 + \log_{2}(\sqrt{wh} / 224)
\end{aligned}
\end{equation} 

The experiment results (denoted as AR$_{FPN}$) are shown in Table \ref{tab:BP4D_control_expr} and Table \ref{tab:DISFA_control_expr}. The AR$_{FPN}$ performs worse than the single-scale RoI feature counterpart AR$_{res}$. This is because AU R-CNN needs high-level RoI features to classify AUs well and does not need to perform box coordinate regression. Furthermore, the bounding boxes in AU R-CNN are not too small to detect compared with those in the object detection scenario. Therefore, pyramidal features are not necessary in detection.

\subsection{Static versus Dynamic}
\label{sec:static_vs_dynamic}
 Can the previous state of facial expression action always improve AU detection? In this section, we conduct a series of experiments using the most popular dynamic models that are integrated into AU R-CNN, including AR$_{ConvLSTM}$, as described in Section \ref{sec:method_AR_conv-lstm}, to determine the answer.
\begin{figure}
	\label{fig:BP4D_correlation}
	\setlength{\abovecaptionskip}{0pt}
	\setlength{\belowcaptionskip}{-0pt}
	\centering
	\includegraphics[width=0.4\textwidth]{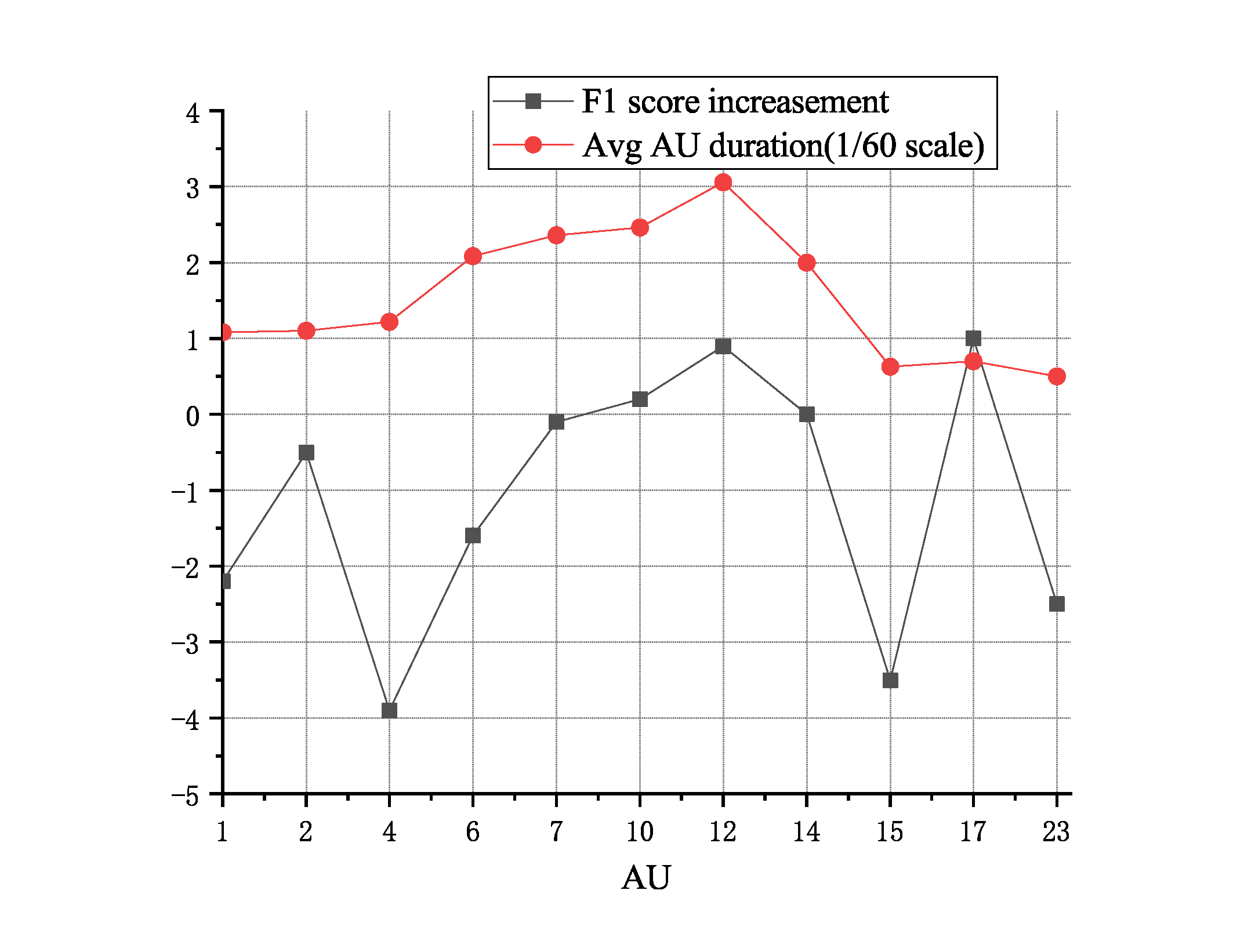}
	\caption{Correlation between F1 score improvement of that in AR$_{ConvLSTM}$ over that in AR$_{res}$ and AU activity duration, AU activity duration is rescaled presenting clarity.}
	\label{fig:F1_duration_correlation}
	
\end{figure}
\subsubsection{AU R-CNN + ConvLSTM}
\begin{table}[!htp]
	\scriptsize	
	\caption{AU average duration \& segments count in \textbf{BP4D}}
	\label{tab:BP4D_AU_DURATION}
	\centering
	\tabcolsep=0.05cm
	\begin{tabular}{c*{12}{p{4.5ex}}}
		\toprule
		AU &  1 & 2 & 4 & 6 & 7 & 10 & 12 & 14 & 15 & 17 & 23 & 24 \\
		\midrule
		Avg duration & 65 & 66 & 73 & 125 & 142 & 148 & 184 & 120 & \textbf{38} & 42 & \textbf{30} & 49 \\
		Seg count & 474 & 380 & 408 & 540 & 569 & 591 & 448 & 571 & 647 & 1203 & 806 & 458 \\
		
		\bottomrule
	\end{tabular}

\end{table}

\begin{table}[!htp]
	\scriptsize
	
	\setlength{\abovecaptionskip}{0pt} 
	
	\caption{AU average duration \& segments count in \textbf{DISFA}}
	\label{tab:DISFA_AU_DURATION}
	\centering
	\tabcolsep=0.05cm
	\begin{tabular}{c*{8}{p{5ex}}}
		\toprule
		AU &  1 & 2 & 4 & 6 & 9 & 12 & 25 & 26 \\
		\midrule
		Avg duration & \textbf{55} & 68 & 112 & 115 & 96 & 133 & 154 & \textbf{82} \\
		Seg count & 320 & 218 & 438 & 340 & 148 & 464 & 600 & 606 \\
		\bottomrule
	\end{tabular}
\end{table}
In this section, we conduct experiments on AR$_{ConvLSTM}$, whose architecture is described in Section \ref{sec:method_AR_conv-lstm}.
Table \ref{tab:BP4D_control_expr} and Table \ref{tab:DISFA_control_expr} present that AR$_{ConvLSTM}$ has a slightly lower average F1 score than AR$_{res}$.  The main reason of the overall performance drop is that the action duration varies drastically in different AUs (Table \ref{tab:BP4D_AU_DURATION} and Table \ref{tab:DISFA_AU_DURATION}); if the temporal length of AU duration is short, ConvLSTM model does not have sufficient capability to observe such actions. The switch of action is so rapid that ConvLSTM cannot infer such label change when processing in the video. We draw a plot of F1 score improvement of AR$_{ConvLSTM}$ over AR$_{res}$ and average AU duration (rescale to 1/60 scale) in Fig. \ref{fig:F1_duration_correlation} to justify our hypothesis. Other factors also influence the performance of ConvLSTM, we can see the red line and the black line have strong correlation in most AUs except AU 1, 2, 4 and AU 15, 17, 23. The reason of high F1 score improvement in AU 17 is that AU 17 has much more segment count (1203) than AU 15 and AU 23 (Table \ref{tab:BP4D_AU_DURATION}), which results in sufficient training samples of AU 17. The AU 4 has lower F1 score improvement than that of AU 1, 2, because AU 4's bounding box (corresponding AU group \#2) is double the size of the area of AU 1 and AU 2 (Fig. \ref{fig:AU_bounding_box}), the larger bounding box leads to weaker recognition capability of capturing the subtle skin change between eyebrows. Most AUs do not have long activity duration; hence, AR$_{res}$ surpasses AR$_{ConvLSTM}$ in average F1 score.

\begin{table*}[t!]
	\scriptsize	
	\caption{The features and applications of dynamic models extension}
	\label{tab:dynamic_summary}
	\centering
	\tabcolsep=0.15cm
	
	\begin{tabular}{llll}
		\toprule
		Model & Application & Training speed & Feature  \\
		\midrule
		AR$_{res}$ & Most cases, no need for the video context & Fast & High accuracy and universal application \\
		AR$_{FPN}$ & Inappropriate &  Medium & Low accuracy and has more layers than AR$_{res}$  \\
		AR$_{ConvLSTM}$ & Suitable for long AU activity duration case & Slow & Accuracy can be improved in long duration activities\\
		AR$_{2stream}$ & Suitable for AUs in small sub-regions especially eye or mouth area & Fast but need pre-compute optical flow & Need pre-compute optical flow first \\
		AR$_{CRF}$ & Application in the case of CPU only & Medium and need pre-computed features & Small model parameter size and no need to use GPU \\
		AR$_{TAL}$ & Inappropriate & Fast & The training cannot fully converged \\
		\bottomrule
	\end{tabular}
\end{table*}

\subsubsection{AU R-CNN + Two-Stream Network}
\begin{figure}[htbp]
	\setlength{\abovecaptionskip}{0pt}
	\setlength{\belowcaptionskip}{-0pt}
	\begin{center}
		\includegraphics[scale=0.28]{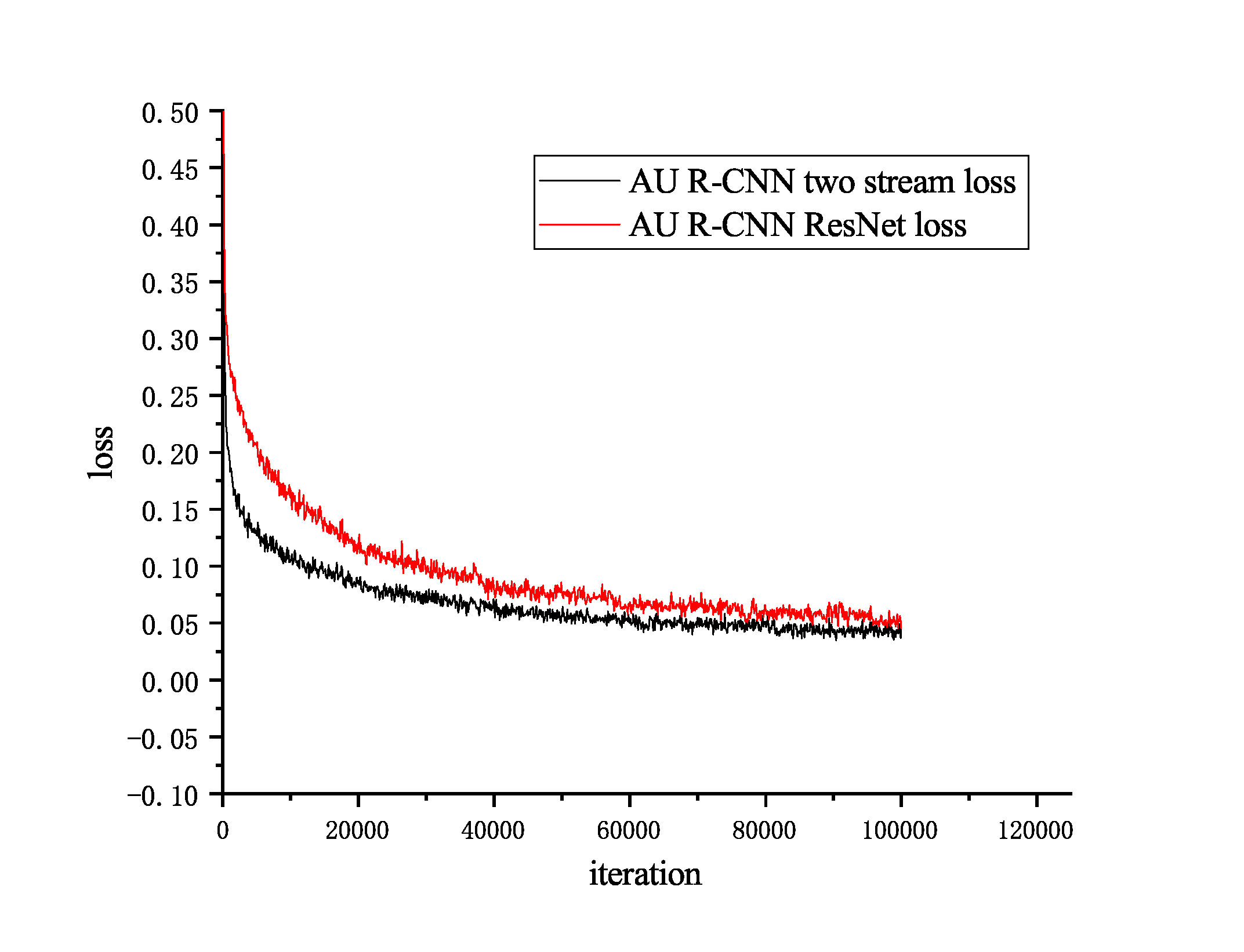}
	\end{center}
	\caption{AR$_{res}$ vs. AR$_{2stream}$ train loss curve}
	\label{fig:two_stream_loss}
	\vspace{-0.0cm}
\end{figure}

Convolutional two-stream network \cite{feichtenhofer2016convolutional} achieves impressive results in video action recognition. In this experiment, we experiment a two-stream network integrated into the AU R-CNN architecture for comparison, denoted as ``AR$_{2stream}$".
We use a 10-frame optical flow and a single corresponding RGB image,\footnote{This corresponding RGB image is in the corresponding location that centers on 10 flow frames} which are fed into two AU R-CNNs. Both AU R-CNN branches use the same bounding boxes, which is the corresponding bounding boxes of RGB image branch, for classification. Two produced $7\times 7\times 2048$ RoI features are concatenated along the channel dimension. The channel size of 4096 feature map is yielded, which will be reduced to 2048 channels using one kernel size of 1 conv-layer. The features are sent to two fc layers to obtain the classification scores. The ground truth label involved in calculating the loss function adopts the single RGB image's labels. 

The performance of the two-stream network AR$_{2stream}$ is remarkably close to that of RGB-image-based AR$_{res}$, which is slightly worse in the BP4D database (Table \ref{tab:BP4D_control_expr}) and is better in the DISFA database (Table \ref{tab:DISFA_control_expr}). In BP4D, the score significantly increases in AU 17 and AU 24 in AR$_{res}$. All these AUs are in the lip area. We attribute this result to the relative small area in the lip area causes the optical flow to be an obvious signal to classify. If we check the result in DISFA dataset in Table \ref{tab:DISFA_control_expr}, this reason can be verified --- AU 1, AU 6, and AU 9 in the DISFA dataset have the smallest AU group areas (Table \ref{tab:DISFA_AREA}), and the F1 scores of these AUs increase. However, the performance of AR$_{ConvLSTM}$ in these AUs cannot be improved compared with AR$_{2stream}$. This justifies that AU group bounding box area is not the reason of the performance improvement in AR$_{ConvLSTM}$ but is the reason of performance improvement in AR$_{2stream}$.
Although the average F1 score of AR$_{2stream}$ is worse than that of AR$_{res}$ in the BP4D database, an interesting property exists in AR$_{2stream}$--- the training convergence speed is faster than that in AR$_{res}$ (see loss curve comparison in Fig. \ref{fig:two_stream_loss}).

\subsubsection{AU R-CNN + TAL-Net}

TAL-Net \cite{chao2018rethinking} follows the Faster R-CNN detection paradigm for temporal action location, and its goal is to detect 1D temporal segments in the time axis of videos. In this experiment, we regard the video sequence as separate segments, and each segment has one label with it.
We use the same RoI parallel line stream with the AR$_{ConvLSTM}$ because we want to detect each region's activity temporal segments. We reformulate the labels of segments in the AU video sequence as a label inside a start and end time interval. In TAL-Net, we use pre-computed AR$_{2stream}$ features. We stack 10 1-D $3\times 3$ kernel conv-layer on the 1-D feature map in the segment proposal network module to generate segment proposals, and we directly feed the pre-computed 1-D feature map into the SoI pooling layer and subsequent fc layers. This network is denoted as ``AR$_{TAL}$".

In our experiment, we determine that AR$_{TAL}$ cannot converge easily, and the loss can only decrease to approximately 1.3 at most (starting from approximately 2.7), which causes AR$_{TAL}$ to perform worse than AR$_{ConvLSTM}$ (Table \ref{tab:BP4D_control_expr}). We can attribute this result to two reasons. First, facial expression is more subtle than the obvious human body action, and the temporal action localization mechanism cannot work efficiently. Second, training 1-D conv-layer and fc layers requires millions of data samples, which cannot be satisfied when converting an entire video to a 1-D feature map. Therefore, this model has the worst performance among all dynamic models.

\subsubsection{AU R-CNN + General Graph CRF}
CRF model is a classical model for graph inference. We experiment with an interesting idea that involves connecting all separate parts of faces in a video to construct a spatio-temporal graph and then using the general graph CRF to learn from such a graph. This model is denoted as ``AR$_{CRF}$". We not only connect RoIs with the same AU group number in the adjacent frames of the time axis but also fully connect different RoIs inside each frame, thereby yielding a spatio-temporal graph. In this method, the entire facial expression video is converted to a spatio-temporal graph using pre-computed 2048-D features extracted by AR$_{res}$ (average pooling layer's output). This graph encodes not only the temporal dependencies of RoIs but also the spatial dependencies of each frame's RoIs. Table \ref{tab:BP4D_control_expr} and Table \ref{tab:DISFA_control_expr} present that AR$_{CRF}$ has a lower score than does AR$_{res}$ in BP4D and DISFA. We attribute this score decrease to the number of weight parameters. In AR$_{CRF}$, we have only $|\mathcal{F}| \times |\mathcal{Y}| + |\mathcal{E}| \times |\mathcal{Y}|^ 2$ weight parameters in total (in BP4D, it is \num{45540}), where $|\mathcal{F}|$ denotes the feature dimension, $|\mathcal{Y}|$ denotes the class number, and $|\mathcal{E}|$ denotes the number of edge type. We extract 2048-D features from the average pooling layer. Two other fc layers exist on top of the average pooling layer in AR$_{res}$, and their weight matrices are $2048 \times 1000$ and $1000 \times 22$, which result in \num{2070000} parameters that are much more than \num{45540} in AR$_{CRF}$. Therefore, classification performance is influenced not only by correlation but also by model capacity (including the number of learned parameters).

\subsubsection{Dynamic models summary}
After above discussion, the features and application of dynamic models extension can be summarized in Table \ref{tab:dynamic_summary}.

\section{Conclusion}

In this paper, we present AU R-CNN for AU detection. It focuses on adaptive regional learning using expert prior knowledge, whose introduction provides accurate supervised information and fine-grained guidance for detection. Complete comparison experiments are conducted, and the results show that the presented model outperforms state-of-the-art approaches and the conventional CNN baseline model which uses the same backbone, proving the benefit of introducing the expert prior knowledge. We also investigate various dynamic architectures that are integrated into AU R-CNN, which demonstrate that the static-image-based AU R-CNN outperforms all the dynamic models. Experiments conducted on the BP4D and DISFA databases manifest the effectiveness of our approach.

\section*{Acknowledgement}
This research is partially supported by the National Key R\&D Program of China (Grant No. 2017YFB1304301) and National Natural Science Foundation of China (Grant Nos. 61572274, 61672307, 61272225).

\bibliographystyle{elsarticle-num}
\bibliography{main}
\newpage

\begin{table}[htp!]
	\scriptsize
	\begin{tabular}{cp{6cm}}
		\raisebox{-0.9\height}{\includegraphics[width=0.15\textwidth]{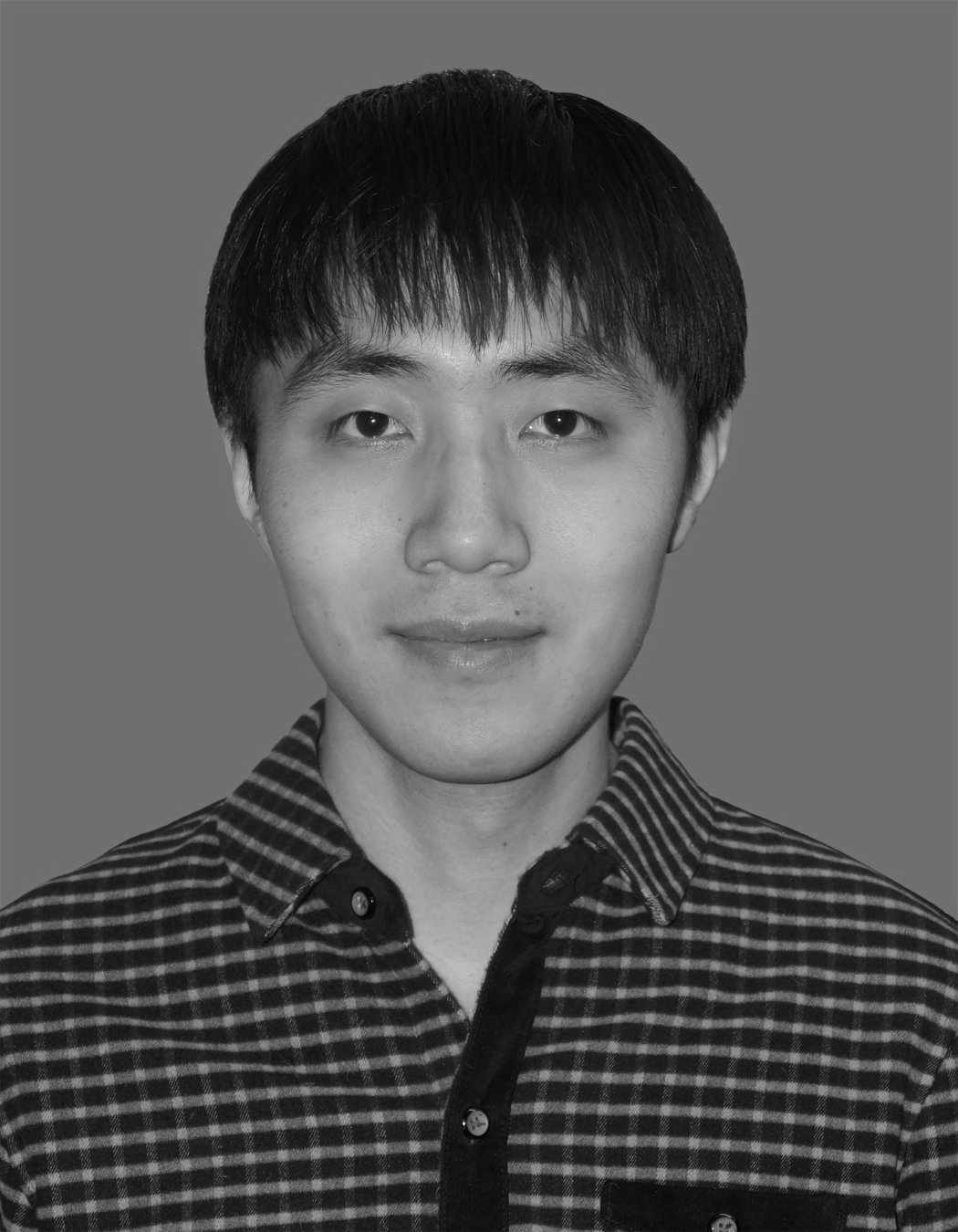}} & \textbf{Chen Ma} is currently pursuing the Ph.D. degree with the School of Software Department, Tsinghua University, Beijing, China. He received the Master degree from Beijing University of Posts and Telecommunications, in 2015. His research interests include facial expression analysis, action unit detection, and deep learning interpretability. \\
		& \\
		\raisebox{-0.9\height}{\includegraphics[width=0.15\textwidth]{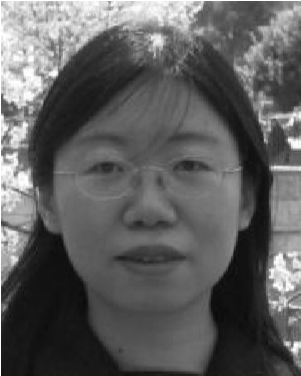}} & \textbf{Li Chen} received the Ph.D. degree in visualization from Zhejiang University, Hangzhou, China, in 1996. She is currently an Associate Professor with the Institute of Computer Graphics and Computer Aided Design, School of Software, Tsinghua University, Beijing, China. Her research interests include data visualization, mesh generation, and parallel algorithm. \\
		& \\
		\raisebox{-0.9\height}{\includegraphics [width=0.15\textwidth]{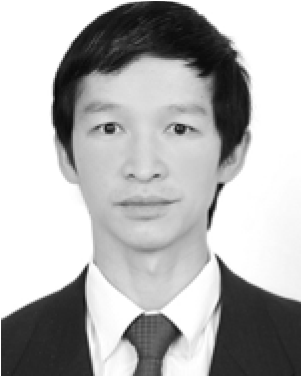}}
		&
		\textbf{Jun-Hai Yong} is currently a Professor with the School of Software, Tsinghua University, Beijing, China, where he received the B.S. and Ph.D. degrees in computer science, in 1996 and 2001, respectively. He held a visiting researcher position with the Department of Computer Science, Hong Kong University of Science and Technology, Hong Kong, in 2000. He was a Post-Doctoral Fellow with the Department of Computer Science, University of Kentucky, Lexington, KY, USA, from 2000 to 2002.
		He received several awards, such as the National Excellent Doctoral Dissertation Award, the National Science Fund for Distinguished Young Scholars, the Best Paper Award of the ACM SIGGRAPH/Eurographics Symposium on Computer Animation, the Outstanding Service Award as an Associate Editor of the Computers and Graphics (Elsevier) journal, and several National Excellent Textbook Awards. His main research interests include computer-aided design and computer graphics. \\

	\end{tabular}

\end{table}

\end{document}